\documentclass[sigconf,natbib=true]{acmart}
\AtBeginDocument{%
  \providecommand\BibTeX{{%
    \normalfont B\kern-0.5em{\scshape i\kern-0.25em b}\kern-0.8em\TeX}}}

\newcommand{\systemname}{\textit{LiNR}}
\usepackage{enumitem} % QQ added
\usepackage{multirow}
\usepackage{adjustbox}
\usepackage{array}
\newcolumntype{R}[2]{%
    >{\adjustbox{angle=#1,lap=\width-(#2)}\bgroup}%
    l%
    <{\egroup}%
}

\usepackage{tablefootnote}
\usepackage{varwidth}
\usepackage{stmaryrd}
\DeclareCaptionFormat{myformat}{%
  \begin{varwidth}{\linewidth}%
    \centering
    #1#2#3%
  \end{varwidth}%
}
\def\position{\text{position}}
\copyrightyear{2024}
\acmYear{2024}
\setcopyright{acmlicensed}\acmConference[CIKM '24]{Proceedings of the 33rd ACM International Conference on Information and Knowledge Management}{October 21--25, 2024}{Boise, ID, USA}
\acmBooktitle{Proceedings of the 33rd ACM International Conference on Information and Knowledge Management (CIKM '24), October 21--25, 2024, Boise, ID, USA}
\acmDOI{10.1145/3627673.3680091}
\acmISBN{979-8-4007-0436-9/24/10}

\newcounter{def_count}
\begin{document}
%\fancyhead{}
%\title{LiRAR: Retrieval as Ranking on GPUs at LinkedIn}
\title{LiNR: Model Based Neural Retrieval on GPUs at LinkedIn}

\author{Fedor Borisyuk}
\authornote{Authors contributed equally to this research.}
\author{Qingquan Song}
\authornotemark[1]
\author{Mingzhou Zhou}
\authornotemark[1]
\author{Ganesh Parameswaran}
\authornotemark[1]
\affiliation{%\small
 \institution{LinkedIn}
 \city{Mountain View}
 \state{CA}
 \country{USA}}
\email{fedorvb@gmail.com}

\author{Madhu Arun}
\author{Siva Popuri}
\author{Tugrul Bingol}
\affiliation{%\small
 \institution{LinkedIn}
 \city{Mountain View}
 \state{CA}
 \country{USA}}
\email{maarun@linkedin.com}

\author{Zhuotao Pei}
\author{Kuang-Hsuan Lee}
\author{Lu Zheng}
\affiliation{%
 \institution{LinkedIn}
 \city{Mountain View}
 \state{CA}
 \country{USA}}
\email{zpei@linkedin.com}

\author{Qizhan Shao}
\author{Ali Naqvi}
\author{Sen Zhou}
\author{Aman Gupta}
\affiliation{%
 \institution{LinkedIn}
 \city{Mountain View}
 \state{CA}
 \country{USA}}
\email{hshao@linkedin.com}

%\author{Fedor Borisyuk, Qingquan Song, Mingzhou Zhou, Ganesh Parameswaran, Madhu Arun, Siva Popuri, \\ Tugrul Bingol, Zhuotao Pei, Kuang-Hsuan Lee, Lu Zheng, Qizhan Shao, Ali Naqvi, Sen Zhou, Aman Gupta}
%\affiliation{\institution{LinkedIn Inc.}}

\renewcommand{\shortauthors}{Fedor Borisyuk et al.}
%% If needed use a foot or authornote to identify equal contribution

\begin{abstract}
%In  this  paper, we  present {\systemname}, a deployed large-scale GPU based retrieval system at LinkedIn. {\systemname} enables the storing of a billion-sized index on the GPU model. We share our insight on developing scalable implementation of differentiable search index using TensorFlow and PyTorch, where items and model weights are stored jointly within the model binary. As index build is considered as model training, we explain how we scaled our system to large indexes with full scan and effective filtering. In particular, we share how to enable attribute based pre-filtering with exhaustive search on GPU, which has been a challenge because many infra systems perform KNN with post filtering degrading system quality. We also describe set of modeling algorithms to enable multi-embedding retrieval and how to address cold start during retrieval. We share how we pushed our system to support larger indexes with quantization. We believe that our ML infrastructure enabled one of the first Live-updated model based retrieval index in the industry. We deployed our solution to out-of-network posts recommendations for the LinkedIn Feed. The techniques presented in this work have contributed a 3\% improvement in professional daily active members. We believe that {\systemname} paves the way for unified retrieval and ranking stack into a single GPU model simplifying complex infrastructure systems and providing ability to optimize whole  differentiable infrastructure end-to-end with gradient descent.

This paper introduces {\systemname}, LinkedIn's large-scale, GPU-based retrieval system. {\systemname} supports a billion-sized index on GPU models. We discuss our experiences and challenges in creating scalable, differentiable search indexes using TensorFlow and PyTorch at production scale. In {\systemname}, both items and model weights are integrated into the model binary. Viewing index construction as a form of model training, we describe scaling our system for large indexes, incorporating full scans and efficient filtering. A key focus is on enabling attribute-based pre-filtering for exhaustive GPU searches, addressing the common challenge of post-filtering in KNN searches that often reduces system quality. We further provide multi-embedding retrieval algorithms and strategies for tackling cold start issues in retrieval. Our advancements in supporting larger indexes through quantization are also discussed. We believe {\systemname} represents one of the industry's first Live-updated model-based retrieval indexes. Applied to out-of-network post recommendations on LinkedIn Feed, {\systemname} has contributed to a 3\% relative increase in professional daily active users. We envisage {\systemname} as a step towards integrating retrieval and ranking into a single GPU model, simplifying complex infrastructures and enabling end-to-end optimization of the entire differentiable infrastructure through gradient descent.

%We believe that this work can provide practical solutions and insights for engineers who are interested in applying retrieval solutions to empower recommendation and search systems that operate at large production scale. 

\end{abstract}

\begin{CCSXML}
<ccs2012>
<concept>
<concept_id>10002951.10003317.10003338.10003342</concept_id>
<concept_desc>Information systems~Similarity measures</concept_desc>
<concept_significance>500</concept_significance>
</concept>
<concept>
<concept_id>10002951.10003317.10003365.10003366</concept_id>
<concept_desc>Information systems~Search engine indexing</concept_desc>
<concept_significance>500</concept_significance>
</concept>
<concept>
<concept_id>10002951.10003317.10003338.10003343</concept_id>
<concept_desc>Information systems~Learning to rank</concept_desc>
<concept_significance>500</concept_significance>
</concept>
<concept>
%<concept_id>10010147.10010257.10010321</concept_id>
%<concept_desc>Computing methodologies~Machine learning algorithms</concept_desc>
%<concept_significance>500</concept_significance>
%</concept>
</ccs2012>
\end{CCSXML}

\ccsdesc[500]{Information systems~Similarity measures}
\ccsdesc[500]{Information systems~Search engine indexing}
\ccsdesc[500]{Information systems~Learning to rank}
%\ccsdesc[500]{Computing methodologies~Machine learning algorithms}

\keywords{information retrieval, recommender systems, candidate generation, nearest neighbor search, neural retrieval}

\maketitle

\section{Introduction}\label{sec:intro}

LinkedIn, the world's largest professional network, serves over a billion members globally, offering services from job searches to content engagement. This paper explores {\systemname}, LinkedIn's model-based GPU retrieval system, focusing on embedding-based retrieval (EBR). Traditional EBR uses unsupervised nearest neighbor search solutions~\cite{hnsw_paper, pq_paper}, indexing item vectors for fast retrieval. Our paper presents an innovative approach, combining exhaustive search with pre-filtering in a differentiable GPU model, using neural networks for distance learning and ranking. In {\systemname}, item vectors and model weights coexist within the same model binary, unlike traditional search indexing methods.

We believe the future of search and recommender systems lies in differentiable model-based serving, enabling joint optimization of retrieval and ranking. The K-nearest neighbor (KNN) search algorithm, an essential embedding-based retrieval method, uses learned query and item embeddings with a specific similarity metric to select the top-K closest items. Typically, KNN uses dot-product similarity, a form of matrix multiplication with normalized embeddings, which has been significantly sped up on modern GPUs (A100, H100, etc.) in frameworks like PyTorch and TensorFlow. Several challenges motivate us to propose model-based KNN algorithms implemented on GPUs:
%\vspace{-4pt}
\begin{itemize}[leftmargin=*]
    \item Liquidity challenge: Real-time search systems rely on specific attributes to filter relevant items. In job recommendation systems, for example, filters like company names, locations, and skills are essential. Items meeting these conditions must be prioritized to avoid exclusion due to low KNN scores from embeddings alone.
    %\item Liquidity challenge: Real-time search systems depend on definite attributes to filter relevant items per query. For instance, in job recommendation systems, postings include company names, locations, and skill requirements—explicit filters users might specify. Items meeting these query conditions must be prioritized in rankings to prevent exclusion due to low KNN scores from embeddings alone.

    \item Low latency requirement: Reducing retrieval latency and increasing throughput is a constant priority.
    %Low latency requirement: reducing the retrieval latency and increasing the throughput is an enduring topic.

    \item Huge memory cost: As number item embeddings and clauses increase, finding ways to lower memory usage and boost computational speed without compromising retrieval quality presents a significant challenge.

    \item Freshness: Demonstrate that model-based approaches can enhance traditional nearest neighbor searches in quality and latency while supporting functionalities like live updates.
\end{itemize}

In this paper we discuss deployment of large-scale, neural model-based retrieval system, highlighting key challenges and solutions. A major challenge was the absence of efficient pre-filtering in PyTorch and TensorFlow, addressed by our custom indexing and filtering methods detailed in \S\ref{sec:knn}, which also tackle latency issues. We also cover memory cost management through quantization techniques for larger indexes in \S\ref{sec:quant_knn}. {\systemname} enhanced search quality, utilizing multi-embedding retrieval algorithms discussed in \S\ref{sec:mexture_similarities}. Our work positions us among the pioneers in the industry in introducing a retrieval model-based serving infrastructure (\S\ref{sec:MLinfra}), showcasing the capability of such model-based retrieval systems to be effectively live-updated at scale (\S\ref{sec:live_update}). We perform our study of model-based index serving focuses on interest-based recommendations on LinkedIn's Feed, also known as out-of-network (OON) recommendations. These recommendations leverage member profiles and previous interactions with the Feed, enabling LinkedIn members to access highly relevant content. We integrate OON content into various LinkedIn surfaces, like Feed and Notifications, based on predicted user engagement likelihood. The effectiveness of OON recommendations is gauged by member interactions with OON content. We use two-tower neural networks to create embeddings for members and Feed Posts, forming a candidate selection vertical for OON in the Feed through EBR with a differentiable model-based search index. {\systemname} significantly outperforms FAISS-based~\cite{FAISS_paper} retrieval system in OON recommendations. We support full-scan model-based index serving on GPUs with latencies as low as 4 ms, handling indexes from 15 million to a billion entries. This capability, along with modeling enhancements, significantly boosts quality as detailed in \S\ref{sec:A_b_test}.

\section{Related Work}\label{sec:related_work}

Industry focus has predominantly been on approximate neighbor search systems, with FAISS ~\cite{FAISS_paper}, ScaNN ~\cite{scann_google_paper}, SONG ~\cite{song_paper}, RAFT~\cite{RAFT_paper} among notable examples. These support algorithms like HNSW ~\cite{hnsw_paper}, IVFPQ ~\cite{pq_paper}, CAGRA \cite{ootomo2023cagra} on CPU and GPU platforms. Termed model-free, these methods use unsupervised algorithms for partitioning space using existing item embeddings, offering flexibility for any item set. In contrast, our approach employs deep neural networks for a model-based search index, fully operational on GPUs. We integrate item indexes with neural network weights within a PyTorch or TensorFlow model, training during index construction and using the model for retrieval.

Recently with more performance and memory available on GPUs several publications have appeared considering model based nearest neighbor search such as \cite{MoL_paper, tay2022transformer, rajput2023recommender, zhang2023modelenhanced}. Mixture of logits (MoL) \cite{MoL_paper} in its production deployed form implements weighted combination of cosine similarities with neural network gates used to infer per distance component weights. The MoL paper does not provide information on examples of implementation of logits components, and which embedings have been used in production. We extend on top of MoL and introduce practical algorithms on how to learn components of MoL. 
Conversely, research by \cite{tay2022transformer, rajput2023recommender, zhang2023modelenhanced} has explored using transformers and generative techniques for search indexes. Unlike our system, which stores item embeddings directly, these studies create semantic structures through clustering and transformers to generate document IDs.

A lot of research has been focused on representation learning with works representing posts and users in social networks \cite{Que2Search_paper, nxtpost_paper, PinnerSage_paper}. As one of the components in MoL we have used approaches similar to \cite{Que2Search_paper, nxtpost_paper, PinnerSage_paper}, and additionally extended it with approaches for cold start infrequent users using clustering representations.
Several previous works have explored the concept of model live-updates, which we expand on in this paper. These works include Monolith~\cite{liu2022monolith}, PERSIA~\cite{persia_paper}, and XDL~\cite{XDL_paper}. In contrast, traditional search engines, as seen in Facebook Search EBR ~\cite{fb_search_embeddings_paper, Que2Search_paper} via Unicorn~\cite{unicorn_paper}, and Lucene~\cite{Lucene_chen2023endtoend}, have primarily focused on live-update functionality for unsupervised indexing techniques such as ~\cite{FAISS_paper}. To the best of our knowledge, our paper represents one of the pioneering efforts in the realm of retrieval-based techniques for live-updating TensorFlow (TF) or PyTorch model-based retrieval indexes at a large-scale production level, with high QPS demands.

\section{Modeling Technology}\label{sec:overview}

In this section we will describe how we modeled and developed exhaustive embedding-based search on GPU with attribute-based matching. We will provide details on how we scaled our model-based index to billion size on a single GPU with quantization. We extend Mixture of Logits (MoL) \cite{MoL_paper} by automatically training cluster embedding components and experimenting with different gating functions and variety of embedding components.

\subsection{Exhaustive Search with Attribute-Based Matching (ABM)} \label{sec:knn}

Considering the post-filtering (filter after similarity-based retrieval) often suffers from the liquidity issue especially combining with ANN algorithms implemented on GPUs \cite{zhao2022constrained}, we first focus on the KNN-based algorithm with attribute-based pre-filtering and introduce several basic approaches adopted to tackle the above challenges.  Strategies to further improve the algorithm and tackle other online serving challenges including the live update problem will be introduced in \S\ref{sec:system_deployment}.

%{\color{red} Future exploration GPU-based implementation could be ... on ... [ref ping's work ANN filter].}
%{\color{red} QQ: we need a figure here for the two approaches}

\begin{figure}
    \centering
    \vspace{-4pt}
    \begin{adjustbox}{max size={\linewidth}{\textheight}}
        \includegraphics{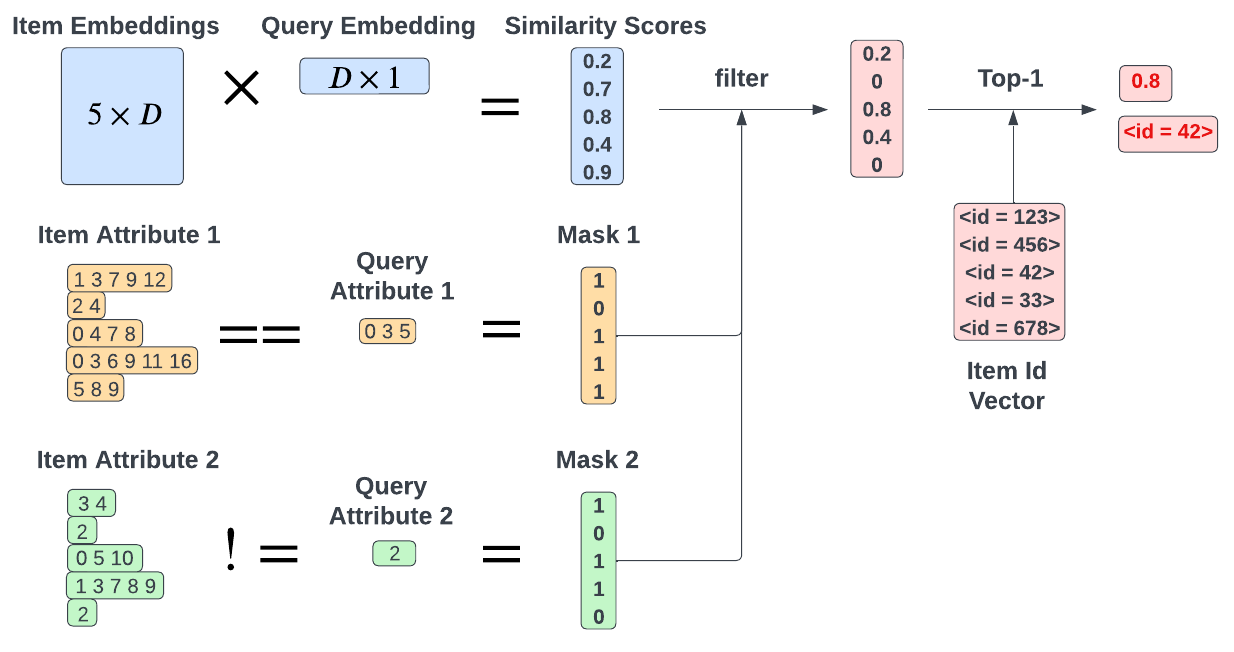}
    \end{adjustbox}
    \caption{
    \small
    %KNN with Similarity Masking: We illustrate with five items and a single query. Item similarities are computed and masked with two 0-1 vectors returned from two clause checking. If any item attribute matches the query attribute, the clause is passed (returns one) in the masking matrix. The second clause is a reverse matching clause. Top-1 selection is used in this example. D represents the dimension of the item embedding.
    KNN with Similarity Masking. An example of five items with single query is used for illustration. Item similarities are computed and masked with two 0-1 vectors returned from the two clause checking. For each item, as long as one attribute is matched with the query attribute, the clause checking is passed (return one) in the masking matrix. The 2nd clause is a reverse matching clause. Top-1 selection is used in this example. D is dimension of item embedding.
    }
    \label{fig:knn-v1}
    \vspace{-16pt}
\end{figure}

\subsubsection{KNN with Similarity Masking}
Our first KNN algorithm with ABM is a two-step similarity masking approach. As shown in Figure~\ref{fig:knn-v1}, given a query, we first compute the similarity between the query embedding with all item embeddings stored in a matrix to capture their semantic relationships in a similarity vector. Then, we filter out irrelevant items by multiplying it with the 0-1 mask vectors given by each clause to map the similarity scores of the filtered items to zero before the top-K selection. Each query clause could contain multiple attributes. Feasible items should satisfy all clauses, requiring at least one of the attribute in each clauses is matched. Reverse clauses are also supported (such as the company name attributes in Figure~\ref{fig:knn-v1}). As each item could contain different number of attributes for each clause, to effectively utilize the GPU memory for saving and update the clauses, we store all clause attributes in a single matrix in practice and have an extra counting matrix to record the number of attributes for each item in each clause similar to the counting matrix in a CSR format but for each item separately without having the indexing vector. Each item clause is sorted before the concatenation for faster judgement (as we can stop checking early as long as one attribute is matched). We implement the algorithm in CUDA and registered the clause filtering kernel as TensorFlow and PyTorch operations to integrate and serve with other modules.

% This version aims to strike a balance between computational efficiency and accuracy in identifying the nearest neighbors.

% This filtering step is crucial for optimizing the subsequent top-k selection process, ensuring that only the most relevant items are considered. 

\begin{figure}
    \centering
    \vspace{-4pt}
    \begin{adjustbox}{max size={\linewidth}{\textheight}}
        \includegraphics{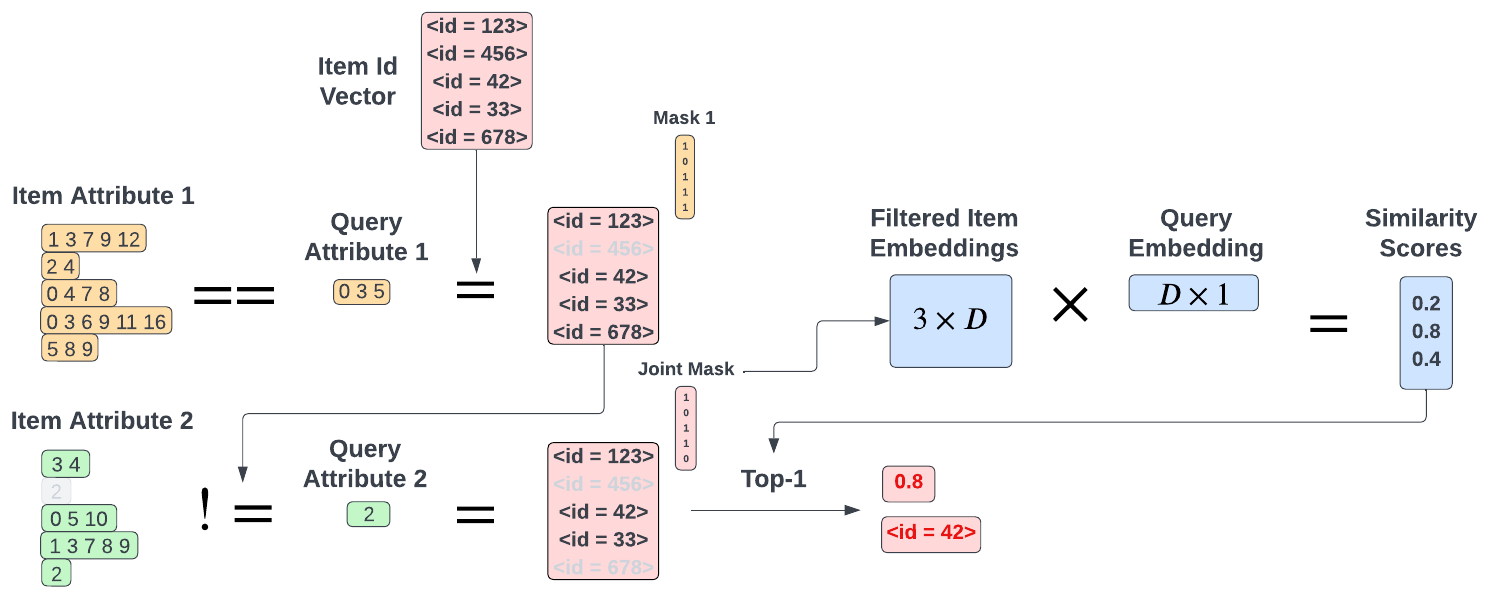}
    \end{adjustbox}
    \caption{\small KNN with Explicit Pre-Filtering. Clauses are checked one by one and a joint 0-1 mask vector is returned to retrieve the feasible items for matrix multiplication and top-K selection (K=1 here).}
    \label{fig:knn-v2}
    \vspace{-12pt}
\end{figure}

\subsubsection{KNN with Explicit Pre-Filtering}\label{sec:prefilter}
The second iteration of our KNN with ABM uses a new approach, incorporating explicit filtering before embedding multiplication, as illustrated in Figure~\ref{fig:knn-v2}. Initially, we slice the matrix to filter out irrelevant items, removing them early from subsequent computations. This method speeds up the process by reducing the computational burden during matrix multiplication and top-K selection, especially beneficial when the query filters result in a significantly smaller item set. We found that with custom CUDA implementation \cite{shen2024learningretrievejobmatching} to merge the kernels, the speed could be generally faster than the first version introduced above. However, without customizing the masked matrix multiplication and kernel merging, simply adopting the matrix slicing in TensorFlow and PyTorch will introduce extra matrix copy and creation overhead, causing it to be slower than the first version when the pass rate is high.
%The second version of our KNN with ABM introduces a different strategy by incorporating explicit filtering before the embedding multiplication. As shown in Figure~\ref{fig:knn-v2}, we begin by filtering out irrelevant items through matrix slicing, and directly remove them from consideration in the subsequent computations. As clause filtering operation is much faster than the rests, this version aims to enhance efficiency by explicitly excluding irrelevant items early in the computation, potentially reducing the computational load during matrix multiplication and top-K selection especially when the passing rate is low during given the query filters, i.e., the refined set of items after the filtering process is much fewer than the original item set. Practically, we found that with good custom CUDA implementation to merge the kernels, the speed could be generally faster than the first version introduced above. However, without customizing the masked matrix multiplication and kernel merging, simply adopting the matrix slicing in TensorFlow and PyTorch will introduce extra matrix copy and creation overhead, causing it to be slower than the first version when the pass rate is high.

% Following this, we perform matrix multiplication to obtain a refined set of similarity scores. 

%The top-k selection process is then applied to identify the most relevant items based on these refined scores. 

\subsection{Quantized KNN}\label{sec:quant_knn}

% \subsection{Index Quantization}
% \todo{QQ to write how to fit large indexes to 1 GPU core}

%{\color{red} QQ: we need another figure here}
%To tackle the memory issue, we introduce a quantized KNN approach leveraging the Sign One Permutation One Random Projection (Sign-OPORP) approach~\cite{li2023oporp} to compress and quantize emeddings into 1-bit values and approximate the dot-product via bitwise-matching operator. This approach provides a trade-off between the prediction accuracy and search speed similar to the regular ANN approach but is an exhaustive search method that can be directly combined with attribute-based pre-filtering method to avoid the liquidity issue.
Addressing memory constraints, we adopt a quantized KNN strategy using the Sign One Permutation One Random Projection (Sign-OPORP) method to compress embeddings to 1-bit and approximate dot-products via bitwise matching. This technique balances prediction accuracy and search speed, akin to typical ANN methods, but as an exhaustive search, it seamlessly integrates with attribute-based pre-filtering, circumventing liquidity issues.
%for fast approximate similarity computation as extra prefilters. 
OPORP is a variant of count-sketch method. It leverages single random projection with fixed-length binning scheme to efficiently project embedding to a low-dimensional embedding. Sign-OPORP takes the sign of the projected embedding to generate 1-bit embedding that could accurately approximate the cosine similarity of the original floating-point embedding~\cite{li2023oporp} via bit-wise matching, i.e., counting the number of matched bits of two quantized 1-bit embedding.

\begin{figure}
    \centering
    \vspace{-4pt}
    \begin{adjustbox}{max size={\linewidth}{\textheight}}
        \includegraphics{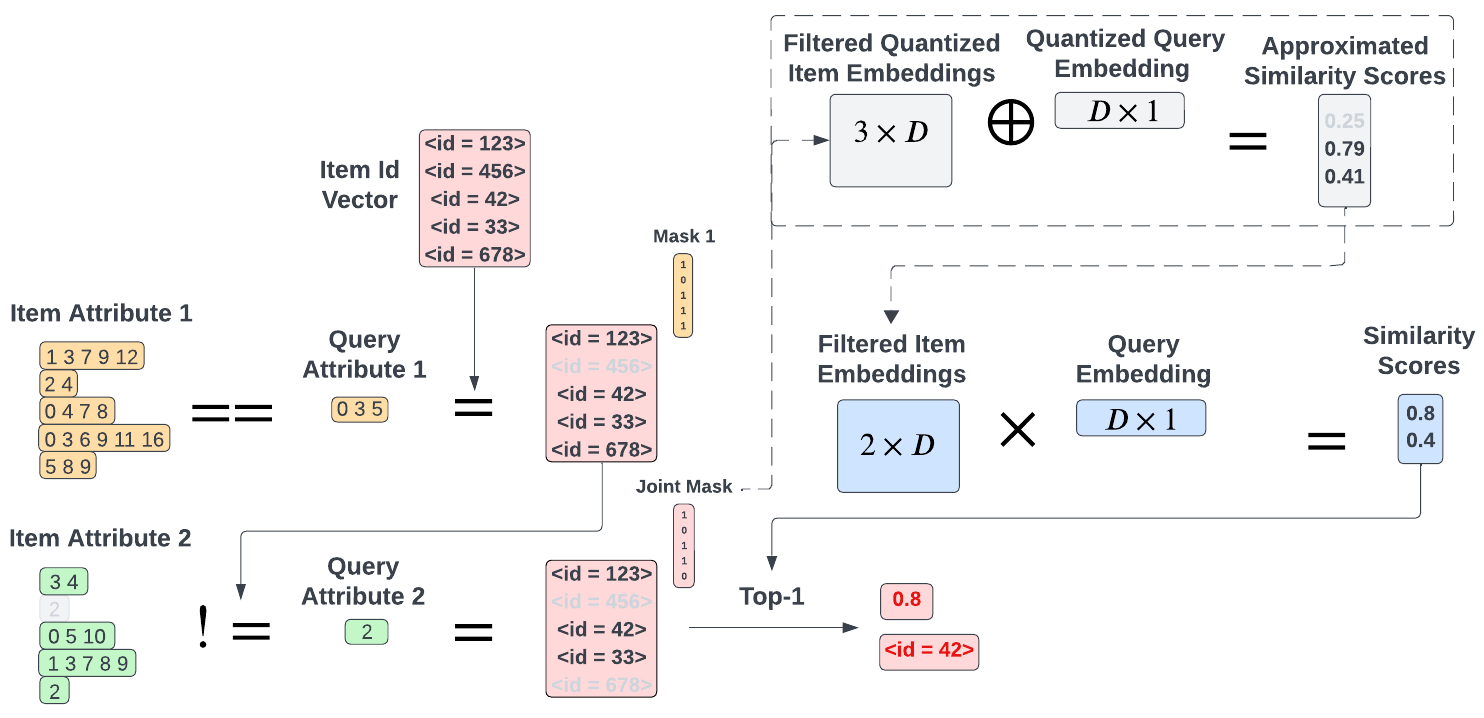}
    \end{adjustbox}
    \caption{\small KNN with Quantized Filtering helps to reduce the number of retrieved items before the full precision similarity computation. A bit-wise matching is used to measure the approximated similarity between 1-bit quantized embedding obtained via Sign-OPORP method. We use bit-wise XOR operation and perform an integer bit-wise NOT conversion for query or item embedding in advance to measure the number of matched bits in the packed integer vector. The quantized KNN module can be used without full precision matrix multiplication when K is large in top-K selection.}
    %The quantized KNN module could also be independently used without having to have the full precision matrix multiplication afterwards when the K value is large in the top-K selection.
    \label{fig:knn-v3}
    \vspace{-12pt}
\end{figure}

As the bit-wise matching operation is often much faster than regular matrix multiplication, we can replace the original embedding with quantized embedding and adopt the bit-wise matching operations in the above-mentioned KNN algorithm with pre-filtering, which can help greatly reduce the memory consumption. Compressing 1 billion fp16 embedding of dimension 64 to 1-bit embedding of the same dimension can reduce the memory by 16 times and help serve 1 billion items in single V100 GPU. We could adjust the size of the quantized embedding to balance the trade-off between the memory/speed and accuracy. Besides, if memory is not the concern, we could leverage the approximated similarity as an extra pre-filtering step reduce the computation of the full-precision matrix multiplication (see Figure~\ref{fig:knn-v3}), offering a unique perspective on exhaustive KNN with ABM. Note that, we call it exhaustive KNN to discriminate it from the regular ANN method with clustering such as HNSW and IVFPQ since our approach still computes all item similarity based on the quantized embeddings, which is easier to be combined with the pre-filtered ABM and live updates (see \S\ref{sec:live_update}).

%This main idea of OPORP is to involves shuffling and quantizing embeddings to 1-bit representations, which are then packed into integers (int32 or int64). Using bit match, we compute an approximated similarity measure of cosine similarity more efficiently. 

%Similar to Version 1, a filtering step is employed, masking the similarity scores to zero for irrelevant items. After selecting the top-T items, matrix multiplication is performed on this refined set, and the final top-K items are selected. The OPORP quantization introduces a trade-off between quantization accuracy and computational efficiency, offering a unique perspective on KNN with ABM.

%{\color{red} Speed is also faster, so if memory issue is not a concern, we can put quantization as an extra filtering step after the clause filtering to further reduce the complexity of computing dot products.}

%% mizhou: change subsection title and add Hadamard MLP
% \subsection{Learnt clustering for Mixture-of-Logits}\label{sec:mexture_similarities}
\subsection{Similarity Modeling}\label{sec:mexture_similarities}

\subsubsection{Hadamard MLP}
Dot Product or cosine similarity has been common in retrieval and it’s computationally efficient. On the other hand, the multilayer perceptron (MLP)-based learned similarity functions has been reported inferior compared to properly tuned dot product. To balance the computation cost/latency and retrieval metrics, we attempted to boost the MLP-based learned similarity function through hadamard product. The architecture is shown on the left of Figure \ref{fig:Residual_learning}. A MLP block is applied to member and item embedding respectively, whose output performs hadamard product and then passes to another MLP block to output the final logit. With proper hyper-parameter tuning, it can reliably outperform dot product.

\subsubsection{Mixture-of-Logits with Clustering}
Mixture-of-logits ~\cite{MoL_paper} defines a model for computing high rank similarity based on adaptive gating of elementary logits across multiple embedding components $\phi_{MoL}(x, u) = \sum\limits_k \pi_{k,\theta}(x, u) \delta_{k,\theta}(x, u)$, where $\pi_{k,\theta}(x, u)$ represents a learnt gating function, which gives per component weight using soft-max gate given input of user and item features. The parameters $\theta$ are learnt through Adam optimization of gradients of a sampled soft-max loss. 
%\begin{equation}
%\small
%$\phi_{MoL}(x, u) = \sum\limits_k \pi_{k,\theta}(x, u) \delta_{k,\theta}(x, u)$
%\end{equation}

\begin{comment}
The authors of MoL used dot product as their similarity function in production settings:
\begin{equation}
\Phi_{\text{MoL dot products}}(x, u) = \sum_k \pi_{k,\theta}(x, u) <f_{k,\theta}(u), g_{k,\theta}(x)> 
\label{eq:MoL} 
\end{equation}
\end{comment}

% $f_{k,\theta}()$ and $g_{k,\theta}()$ are different learnable functions for the raw user and item embeddings. 

\begin{comment}
MoL paper lacks specificity how different MoL components $f_{k,\theta}(u)$ and $g_{k,\theta}(x)$ are trained. To fill the gap we provide variety of methods to generate $f_{k,\theta}()$ and $g_{k,\theta}()$ embedding components for MoL in subsections \S\ref{sec:residual_IDs} and \S\ref{sec:offline_evaluation}.
\end{comment}

\begin{comment}
\cite{MoL_paper} mentions that dot product similarities can outperform multilayer perceptron (MLP)-based learned similarity functions. In contrast we extend the formulation to provide set of learnt distances $F_{i}$ with MLP layers including weighted inner product \cite{weighted_dot_product} represented by a single linear layer and Hadamard MLP learnt with few fully connected layers \cite{wang2022flashlight}:
\begin{equation}
\Phi(x, u) = \sum_{F_{i} \in \text{ MLP similarities}} \sum_k \pi_{k,\theta}(x, u) F_{i}(f_{k,\theta} (u), g_{k,\theta}(x))
\label{eq:mixture_of_similarities}
\end{equation}
\end{comment}

Mixture-of-logits requires the availability of multiple features to leverage the gates, because the gates will collapse to a value of 1 if there is only one feature for user and item pair. We augment the feature with learnt cluster id embedding that obviate the necessity for having multiple features.
In \S\ref{sec:offline_evaluation} we show that learnt cluster id embedding leveraged through Mixture-of-Logits
can significantly improve on top of dot product in production settings.

\begin{figure}
    \centering
    \vspace{-4pt}
    \begin{adjustbox}{max size={\linewidth}{\textheight}}
        \includegraphics[scale=0.1]{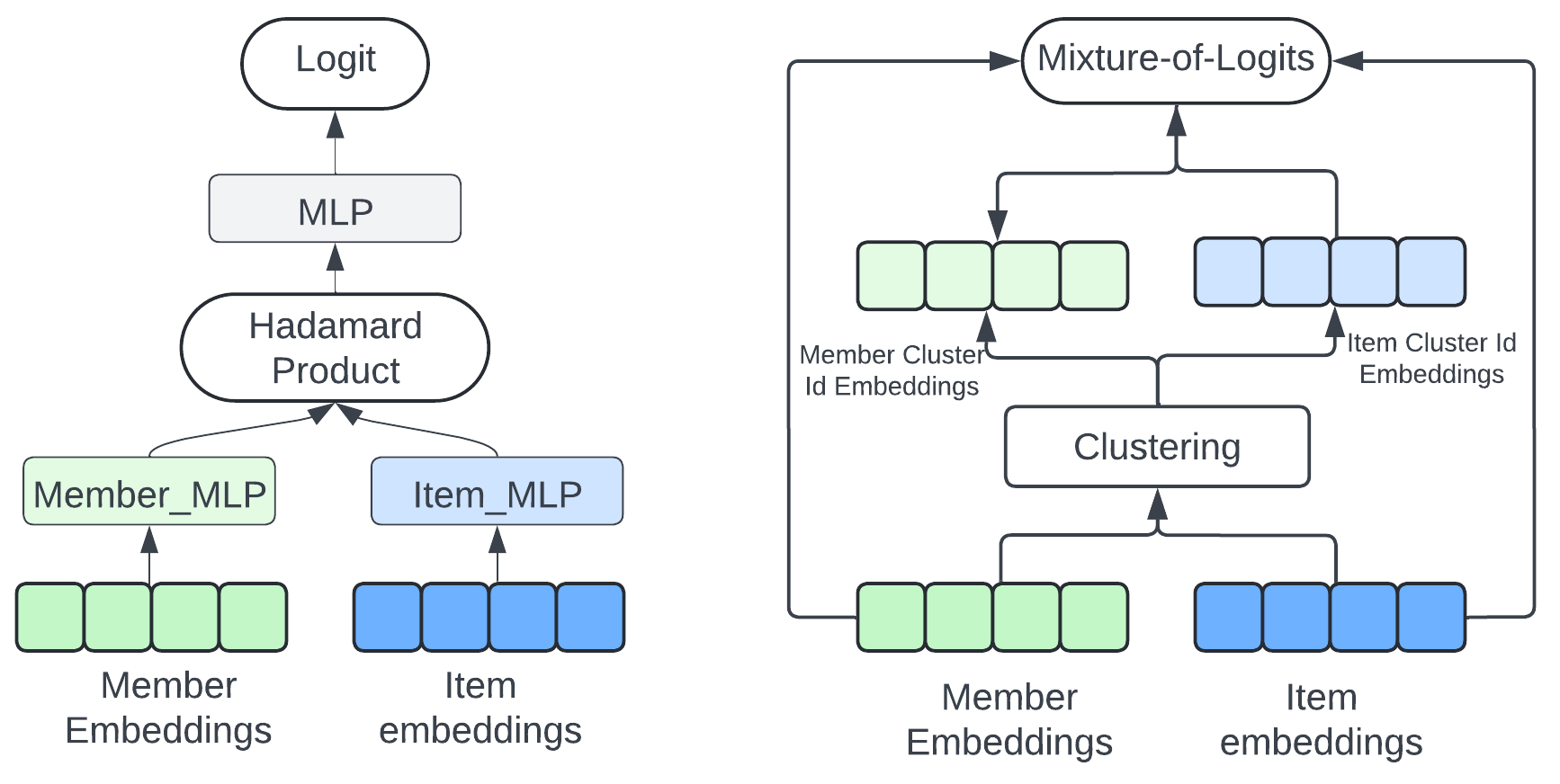}
    \end{adjustbox}
    \caption{\small Illustration of Hadamard MLP (left) and learning cluster id embedding for Mixture-of-Logits(right)}
    \label{fig:Residual_learning}
    \vspace{-12pt}
\end{figure}

% \subsubsection{Learning cluster id embedding for Mixture-of-Logits}\label{sec:residual_IDs}
Across LinkedIn we observed variety of member behaviour with some members coming frequently and some coming from time to time. For the infrequent members we aimed to improve retrieval system performance. To achieve this we learn cluster id embeddings, which represent interests of cohorts of members and topics of posts. We describe the process on the right of Figure \ref{fig:Residual_learning}. 

% Given two tower embeddings of members and posts, we cluster them using Kmeans++ in offline and store cluster centers as part of the {\systemname} model.

For training {\systemname}, we obtain two-tower embeddings for posts and members as part of the training data, along with available engagement labels. We initialize cluster ID embeddings using K-means on millions of post embeddings. During training for both members and posts, we find the closest cluster ID embedding based on cosine similarity to their two-tower embedding. These cluster IDs for members and posts are integrated into Mixture-of-Logits, along with the original two-tower embedding and other embeddings we developed for our use cases. We experimented with using K-means-initialized cluster ID embeddings as is and fine-tuning them through back propagation. We report the experiment results in \S\ref{sec:offline_evaluation}.

\begin{comment}
During training and inference given the member or post embeddings we infer cluster centers associated with it by dot product of post or member two tower embedding with cluster centers. We take top $K$ clusters. For each of the cluster IDs we train respective clusterID embeddings as part of the model based on the member's engagement data, which means that clusters represent engagement patterns of those topics. ClusterID's usage in the model as a sparse ID feature is a common technique \cite{visrel_paper}. In this paper we propose to extend clusterID features and train them jointly with personalized ID embeddings for popular members and posts and sum them together in the model for more robust feature representation. Combination of clusterID with personalized IDs helps to improve learning for cold start items/members and kicks in personalization given even few engagement points for items/members via personalized ID embeddings. Cluster IDs can be seen as initialization for personalized ID embeddings due to sum combination in the model. Member ID and post ID embeddings are randomly initialized closely to zero. ClusterID embeddings could be initialized randomly or with offline kmeans centroid values. We found that performance is the same with longer training for both initialization methods.
\end{comment}

\section{System Architecture}\label{sec:system_deployment}

\subsection{Out-of-Network Recommendations}
%Out of Network Recommendations serve as one of several initial ranking sources for the LinkedIn Homepage Feed. When a member accesses the feed, a request is sent from the front end to the feed service, which then delegates to several first pass rankers, including the feed-OON mid tier (a.k.a. interest discovery), tasked with discovering interests. This service retrieves the top-K most suitable items for the member using a lucene-based index.

%As shown in Figure \ref{fig:OONArchitecture}, each member query triggers an embedding-based search across all eligible item embeddings, followed by a first-layer (L1) ranking model to select the top-K items. These are then forwarded to the feed service and undergo a second, more complex layer (L2) ranking model for member viewing.

%{\systemname} is designed to run online model-based retrieval algorithms, aiming to surpass our existing benchmarks: (1) dot-product based EBR, and (2) FAISS-IVFPQ, both integrated within LinkedIn's lucene systems.

\begin{figure}
  \centering
  \vspace{-4pt}
  \begin{adjustbox}{max size={\linewidth}{\textheight}}
  \includegraphics{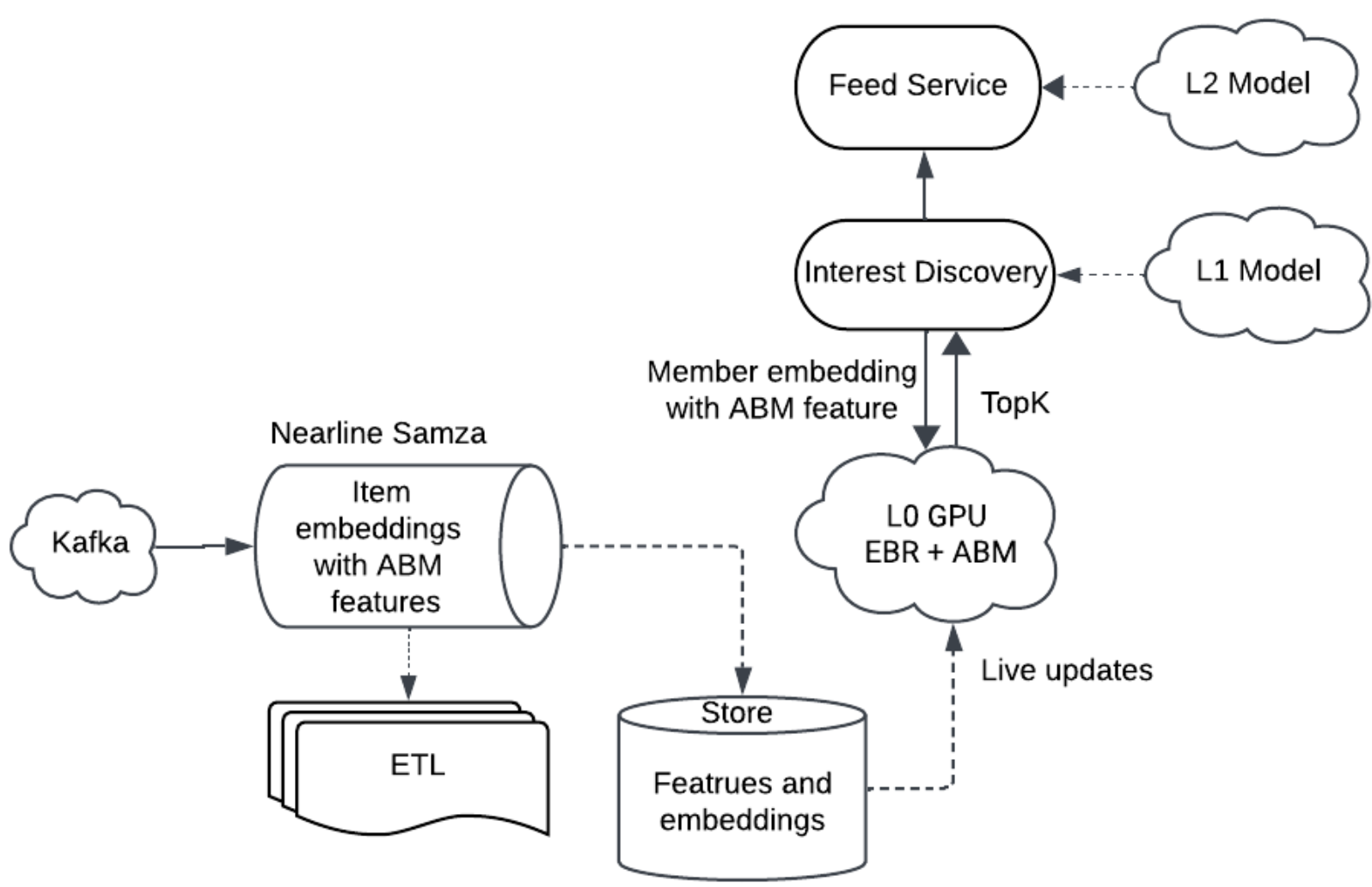}
  \end{adjustbox}
  \caption{\small Feed OON Architechture.}
  \label{fig:OONArchitecture}
  \vspace{-12pt}
\end{figure}

Out of Network Recommendations is one of the many sources (first pass rankers) of Linkedin Homepage Feed. When a member visits Linkedin feed, a request is triggered from the front end and sent to feed service. Feed service passes this request to many first pass rankers including feed-OON mid tier (a.k.a. interest discovery). This service is responsible for retrieving the top-K most eligible items for the member to send back to feed. Today, the underlying index used for retrieval is a lucene based index. 
The runtime of the query of OON is depicted at Figure \ref{fig:OONArchitecture}. For every member query, a embedding based search is performed across all eligible item embeddings, followed by a layer 1 (L1) ranking model, which decides top-K items. 
These are then sent to feed service and ranked by more sophisticated layer 2 (L2) ranking model for members consumption.
{\systemname} aims to provide an online service to run model based retrieval algorithms that can outperform our baselines: (1) dot-product based EBR, and (2) FAISS-IVFPQ, which is supported at Linkedin for lucene systems. 

%Being a new product that was tested on Linkedin feed, OON heavily relies on an offline inference framework to run embedding based retrieval.
%The top-K items are then stored on a cloud based storage for online lookup. During online query, this top-K is retrieved, gets further filtered by lucene term based product rules, 

As shown in the figure, interest-discovery will call model-cloud-L0 to fetch candidate items for the member. Model-cloud-L0 hosts the RAR model that does (1) item attribute-based filtering (2) embedding based retrieval with ranking using model. The model consists of the item embeddings, features needed for filtering and the trained model weights.  Item embeddings are generated on a nearline fashion as and when a document is created at Linkedin so as to keep the index up to date. The filters required for filtering are also ingested nearline.

\subsection{ML Infra Architecture}\label{sec:MLinfra}
We enhance Model Cloud, our hosted solution for serving model inferences, to support retrieval as ranking as shown in Figure 5.
\subsubsection{Retriever}
This component performs attribute-based filtering and embedding-based retrieval of the top-k documents for a query. At startup, retriever initializes with the retrieval model and bootstrapped data. Its framework-agnostic design allows easy extension to any framework, such as Torch or TensorFlow. AI engineers can experiment with new methods by developing and deploying corresponding models to this system.

%This component is responsible for attribute based filtering and embedding based retrieval of the top-k documents for a given query. At startup, retriever initializes itself with the retrieval model and bootstrapped data. Being framework agnostic makes this component easy to extend to any modeling framework such as Torch, TensorFlow. Plug and Play:  AI Engineers can experiment with innovative approaches such as the ones discussed in this paper by simply developing a corresponding model and deploying to this system.
\subsubsection{Ingestor}
Model-based retrieval requires the entire document corpus to reside in GPU memory for low latency. To provide fresh results, this corpus must be updated near real-time (nearline). Several following components work together to achieve this functionality.
%Model based retrieval and its low-latency requires the entire corpus of documents live in GPU memory. To serve fresh results, this corpus of documents must also be updated near real time (nearline). We have the following components working together to achieve this functionality.

\textit{Index Store:}
%Features such as attributes and embeddings are delivered by offline data sources and nearline data streams. We leverage Apache Beam in our feature delivery system to join and transform feature data for the whole corpus of documents. Offline, the entire corpus is batch-pushed to a Venice Store. Nearline, updates to the corpus are written to the same Venice Store.
Attributes and embeddings come from offline sources and nearline data streams. We use Apache Beam to join and transform feature data for the entire document corpus. Offline, the full corpus is batch-pushed to a Venice Store. Nearline updates are also written to this Venice Store.

\textit{Updator:}
Updator subscribes to the Index Store's Change-Data-Capture (CDC) Stream. As the feature data gets batch pushed and live updated, the Updator gets notified to further process them and write to the model.

\textit{Bootstrapper:}
At startup, the Ingestor bootstraps from the Venice CDC client by replaying all data from the beginning. The entire data corpus is transformed into the required format and copied to the GPU. To minimize bootstrapping time, we regularly compact the bootstrap data and store a snapshot on disk for a fast warm start.
%When retrieval service starts up, Ingestor bootstraps from the Venice CDC client by replaying all the data from the beginning of time. With the entire corpus of data in memory, it is transformed into the required format and copied to the GPU. We ensure that the bootstrap data is compacted regularly to keep the bootstrapping time to a minimum. We also store a snapshot on disk to enable fast warm start of the service.
\subsubsection{Service}
To meet our performance needs, we avoid the latency and unpredictability of managed, garbage-collected languages. We also minimize network hops, data copies, and transformations. Our Model Cloud L0 service is written in a native language with minimal data transformations. User queries from the L0 client land directly on our service, ensuring we meet latency requirements.
%To meet the scale and performance requirements of our members, we cannot afford to take on the latency overhead and unpredictability that comes with managed, garbage-collected languages. Additionally, network hops, data copies and data transformations should be kept at a required-minimum. Keeping this in mind, our Model Cloud L0 service is written in a native language with minimum to no data transformations. User's query from the L0 client lands directly on our service. This way we are able to meet the latency requirements of our users.
\begin{figure}
    \centering
    \vspace{-4pt}
    \begin{adjustbox}{max size={\linewidth}{\textheight}}
        \includegraphics{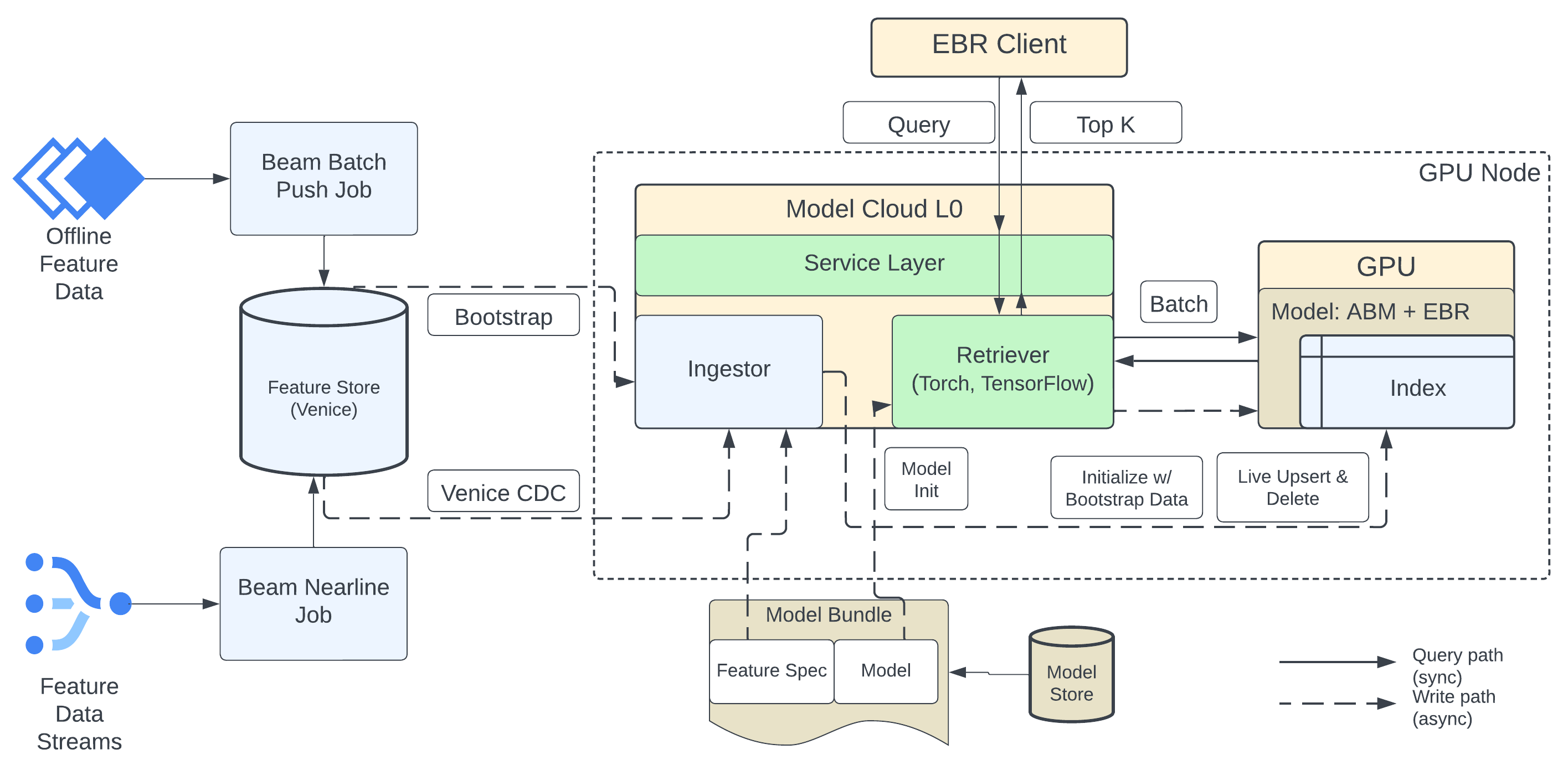}
    \end{adjustbox}
    \caption{\small Model Cloud {\systemname} Architecture}
    \label{fig:enter-label}
    \vspace{-14pt}
\end{figure}

\subsection{Model Live Update}\label{sec:live_update}
Live Update Ingestor subscribes to Venice CDC~\cite{veniceDB} from the bootstrapped offset, classifying changes into upserts and deletes, then transforming and copying them to the GPU.

The system's effectiveness depends on the quality of the document index, which must remain fresh. This can be done by either regularly rebuilding the index or updating it in near-real-time via a data change stream. We chose the latter for two reasons: it keeps the corpus current, reflecting changes within seconds, and it's more efficient, avoiding the cost of rebuilding and replacing the entire index. To implement this, we modified the PyTorch model to expose Upsert and Delete APIs, ensuring safe and efficient concurrent index updates during inference. Techniques used include pre-allocating larger tensors, using a high-water mark to track the working set, and making thread-safe in-memory tensor manipulations with minimal data access serialization. These methods ensure that modifications have minimal to no impact on the inference path, as detailed in the Model Inference Benchmarking section.
%Live Update Ingestor subscribes to Venice CDC from the bootstrapped offset. Changes are classified into upserts and deletes, transformed into the correct format, and copied to the GPU.

%Even though the system functions to surface the best candidates for ranking, its effectiveness is bounded by the quality of the underlying document index. In that aspect, it is essential to keep the index fresh by applying the changes in the underlying data. This could be achieved by either rebuilding the index in regular intervals or constantly updating it in near-real-time with changes pushed through a data change stream. We opted to implement the latter approach as its benefits are twofold. It offers a more current corpus as changes are typically reflected within a few seconds of their creation. Additionally, it is more efficient, as it avoids the cost associated with rebuilding the entire index and replacing a set of older ones. To follow this approach, we modified the PyTorch model to expose Upsert and Delete APIs and implemented them carefully so that the concurrent modifications to the index during the inference can be done safely and efficiently. We used techniques like pre-allocating larger tensors than the initial index requires, using high-water-mark to mark the working set of the tensors and making thread-safe in-memory manipulations on the tensors with minimal data access serializations. These details resulted in modifications having minimal to no impact on the inference path, as described in the Model Inference Benchmarking section.

\subsection{Inference on Native Stack}
We built a native serving system as we made performance and efficiency our top priorities. To serve the PyTorch model in this system, we had to convert it to a compatible format. There are a few alternatives for this purpose such as TorchScipt and torch.export. PyTorch supports two execution modes: eager mode and graph mode. Optimal performance is achieved by executing everything in graph mode as the operators are first synthesized into a graph, which are compiled and executed as a whole. We picked TorchScript for our initial implementation to execute the model in graph mode. However, by doing so we traded off the performance with ease of development. TorchScript is a subset of Python and comes up with some constraints. It requires static typing and does not support things like exceptions and data-dependent control flows. We found executing this conversion quite challenging and concluded that it should be a part of the model development rather than an afterthought. We also decided to pursue other options which are deemed to be more recent technologies such as torch.export.

\section{Experiments}

In this section we provide results on modeling ablation studies, online A/B experiments with OON application and infrastructure model-based retrieval inference benchmarking for production indexes.

\subsection{Model offline evaluation}\label{sec:offline_evaluation}
We evaluated {\systemname} model on our internal dataset. The dataset consists of millions of examples where a member interacted with an item. The member and item are represented by embeddings learnt from a two-tower model.
The two-tower model contains variety of features including member interaction history modeled by \cite{nxtpost_paper} and member profile features, which usually contains member job title, job location, company, skills and professional summary of the member. Posts usually contain text, image, video or external link information. Therefore, such content features from member and posts can help us to identify topics of interests of posts or professional topics of members. 

We used Hit Rate @ 400 over evaluation dataset to report the metrics. We use cosine similarity with exhaustive search as baseline to evaluate against {\systemname}. We report results of Hadamard MLP for single embedding feature and extended Mixture-of-Logits with clustering for both single and multi-embedding features in Table \ref{table:offline_eval}.

Hadamard MLP is favored for production due to its simplicity for deployment and low latency. However, we found Hadamard MLP is very sensitive to weight initialization and general initialization methods such as GlorotNormal or HeNormal can't stabilize the performance. Empirically we observed that the initial few steps determine the overall training trend, Thus we reinitialize the model if the first 100 steps go south.

In addition to the two-tower model and cluster ID features, we enhanced Mixture-of-Logits by introducing multiple embedding features developed at LinkedIn. We incorporated Graph Neural Network (GNN) embeddings for members and posts mapped to the same space using a heterogeneous GNN~\cite{borisyuk2024lignn}. We found that adding more embeddings improved the Hit Rate @ 400 (see Table \ref{table:offline_eval}).

Our extended Mixture-of-Logits with clustering perform well for both single and multi-embedding features. One surprise finding is that fixed clusters (non-trainble) outperform trainable clusters in all cases we explored, one possible explanation is the convergence pace of the clustering and other trainable parameters are different, we'll further investigate it in our future work. Another interesting observation is that it was important to carefully tune the number of clusters: having either too high or too low a value can cause performance to degrade.
\begin{table}[ht]
% centering
\small % This command makes the text of the table smaller
\vspace{-4pt}
\begin{tabular}{lcc}
\hline
\textbf{Method}
& \textbf{Gain in Hit Rate @ 400} \\\hline 
Cosine similarity & --  \\\hline
\textbf{Single Embedding Feature} \\ %\hline
Hadamard(Member \& Item MLP [50]+[10, 1]) & 10.21\% \\
MoL with 70 trained | non-trained clusters & 1.33\% | 10.11\% \\
MoL with 100 trained | non-trained clusters & 11.97\% | \textbf{15.16\%} \\
MoL with 150 trained | non-trained clusters & 4.26\% | 11.17\% \\\hline
\textbf{Multiple Embedding Features} \\ %\hline
MoL without clustering & 12.80\% \\
MoL with 140 trained | non-trained clusters & 20.75\% | 22.61\% \\
MoL with 200 trained | non-trained clusters & 16.49\% | 22.34\% \\
MoL with 300 trained | non-trained clusters & 19.04\% | \textbf{23.67\%} \\\hline

% MoL with 8 clusters & -70.1\%  \\
% \textbf{MoL with 69 clusters} & \textbf{8.56\%} \\
% MoL with 573 clusters & 1.58\%  \\
% MoL with 4723 clusters  & -3.48\% \\\hline
\end{tabular}
\caption{\small Hit Rate @ 400 for single embedding of two-tower model alone in Hadamard or combined with cluster id in MoL, and multiple embeddings combining two-tower, GNN, and cluster ID in MoL.}
\label{table:offline_eval}
\vspace{-16pt}
\end{table}

\subsection{A/B test of \systemname}\label{sec:A_b_test}
For our baseline a dot-product EBR is done across all eligible items given member embedding query. We enabled cache on a cloud based storage for online lookup of computed results. This top K is retrieved by mid tier service when an online feed request is received. 
We leveraged this retrieval framework to test RAR based algorithms to understand relevance impact. The baseline for these experiments are full scan dot-product.

\begin{table}[ht]
\small
\vspace{-4pt}
\begin{tabular}{ll}
\hline
\textbf{Metric Name}
& \textbf{Metric Lift} \\\hline 
{Total professional interactions}            & +7\%   \\
{Daily Unique Gold Professional Interactors}  & +3\%   \\
{Feed Update Views With 30+ Secs Dwell}   & +2\%   \\
{Feed Update Viewers With 30+ Secs Dwell} & +5\%   \\
{Skipped Update Rate}                         & -20\% \\\hline
\end{tabular}
\caption{\small {\systemname} A/B test relative metric improvements.}
\label{table:wdotonline}
\vspace{-12pt}
\end{table}

Table~\ref{table:wdotonline} shows the A/B test results from ramp of {\systemname}. \emph{Total professional interactions} are the total amount of high quality interactions in the form of reshares, reposts, comments, message responses, reacts, votes, saves, and long dwells. \emph{Daily Unique Gold Professional Interactors} is the daily moving average of the number of members or companies generating high quality interactions. \emph{Feed Update Views With 30+ Secs Dwell} counts the total number of feed updates viewed with 30+ secs dwell time. \emph{Feed Update Viewers With 30 Plus Secs Dwell} counts the total number of unique members that viewed a feed update for 30+ secs. \emph{Skipped Update Rate} is the ratio of updates that are skipped (viewed for less than 2 seconds) compared to all viewed updates.

\subsection{Model Inference Benchmarking}\label{sec:model_benchmark}
We conducted offline experiments to benchmark the effectiveness of different framework implementations of two KNN variants with ABM on datasets with different pass-rate scenarios. V1 represents KNN with similarity masking, and V2 represents KNN with explicit pre-filtering. Both are implemented in TF and PyTorch with CUDA kernel for attribute matching registered as a custom operator. 
We selected the Job recommendation index for benchmarking due to its variety of filters, providing multi-dimensional performance insights for {\systemname}. The high-pass-rate and low-pass-rate datasets are derived from job search tasks, containing around 15.5 million jobs with 25 thousand queries. The high-pass-rate dataset includes two clauses: geo-location matching and company name reverse matching (mismatched items are returned), with an average of 1.7 million items passing the clauses. The low-pass-rate dataset includes an additional job title exact matching clause, with a maximum pass rate of 1.2 million for single title matching and most queries having only thousands of passed items. Each item and query has one attribute per clause, converted to 64-bit integers before GPU comparison. The embedding dimension is 128, stored as fp16 values. Performance is measured by average latency, p95 latency in milliseconds, and recall label@2000.

\subsubsection{High-Pass-Rate ABM Dataset Benchmarking}\label{sec:high-pass}

\begin{table}[ht]
\centering
\small
\vspace{-4pt}
%\scriptsize 
\begin{tabular}{@{}lccccc@{}}
\hline
\textbf{Method} & \textbf{Batch} & \textbf{\begin{tabular}[c]{@{}c@{}}Avg. Latency\\ (ms/batch)\end{tabular}}

& \textbf{\begin{tabular}[c]{@{}c@{}}P95 Latency\\ (ms/batch)\end{tabular}}

%\textbf{Avg. Latency (ms/batch)}  
%& \textbf{P95 Latency (ms/batch)} 
& \textbf{Recall@2k} \\ \hline
TF-V1              & 1  & 6.3    & 6.9  & 0.688 \\
TF-V2              & 1  &  6.9   & 14.4 & 0.688 \\
PyTorch-V1         & 1  &  4.8   & 4.9 & 0.688 \\
PyTorch-V2         & 1  &  14.6   & 47.8 & 0.688 \\\hline
% CUDA-V2            & 1  &     &  & 0.688 \\
% CUDA-V3-Q400k      & 1  &     &  & 0.685 \\\hline
% PyTorch-V3-Q400k      & 1 &     &  & 0.684\\\hline

TF-V1              & 16 &  34.8   & 36.6  & 0.688 \\
PyTorch-V1         & 16 &   22.8  & 23.1 & 0.688 \\\hline
% CUDA-V2            & 16 &     &  & 0.688 \\
% CUDA-V3-Q400k      & 16 &     &  & 0.685 \\\hline
% PyTorch-V3-Q400k      & 16 &    &  & 0.684 \\\hline
\end{tabular}
\caption{\small Comparison of implementations on high-pass-rate dataset.}
\label{table:high_pass_knn}
\vspace{-18pt}
\end{table}

% \begin{table}[ht]
% \centering
% \scriptsize % This command makes the text of the table smaller
% \begin{tabular}{lccccc}
% \hline
% \textbf{Method} & \textbf{Batch} & \textbf{Avg. Latency (ms)}  & \textbf{P95 Latency (ms)} & \textbf{Recall@2k} \\ \hline
% TF-V1              & 1  &  16.9   & 17.5  & 0.688 \\
% TF-V2              & 1  &  15.9   & 39.1 & 0.688 \\
% PyTorch-V1         & 1  &  12.8   & 13.0 & 0.688 \\
% PyTorch-V2         & 1  &  12.8   & 13.0 & 0.688 \\
% % CUDA-V2            & 1  & 5.1    & 10.5 & 0.688 \\
% % CUDA-V3-Q400k      & 1  & 3.9    & 5.5  & 0.685 \\\hline
% PyTorch-V3-Q400k      & 1 &  10.5   & 11.0 & 0.684\\\hline

% TF-V1              & 16 &  92.0   & 94.0 & 0.688 \\
% PyTorch-V2         & 16 &  57.6   & 57.8 & 0.688 \\\hline
% % CUDA-V2            & 16 &  30.4   & 39.2 & 0.688 \\
% % CUDA-V3-Q400k      & 16 &  27.2   & 30.7 & 0.685 \\\hline
% PyTorch-V3-Q400k      & 16 &  55.5   & 56.0 & 0.684 \\\hline
% \end{tabular}
% \caption{Comparison of different implementations on high-pass-rate dataset on single V100 GPU.}
% \label{table:high_pass_knn}
% \end{table}

\begin{figure}
    \centering
    \vspace{-4pt}
    \begin{adjustbox}{max size={\linewidth}{\textheight}}
        \includegraphics{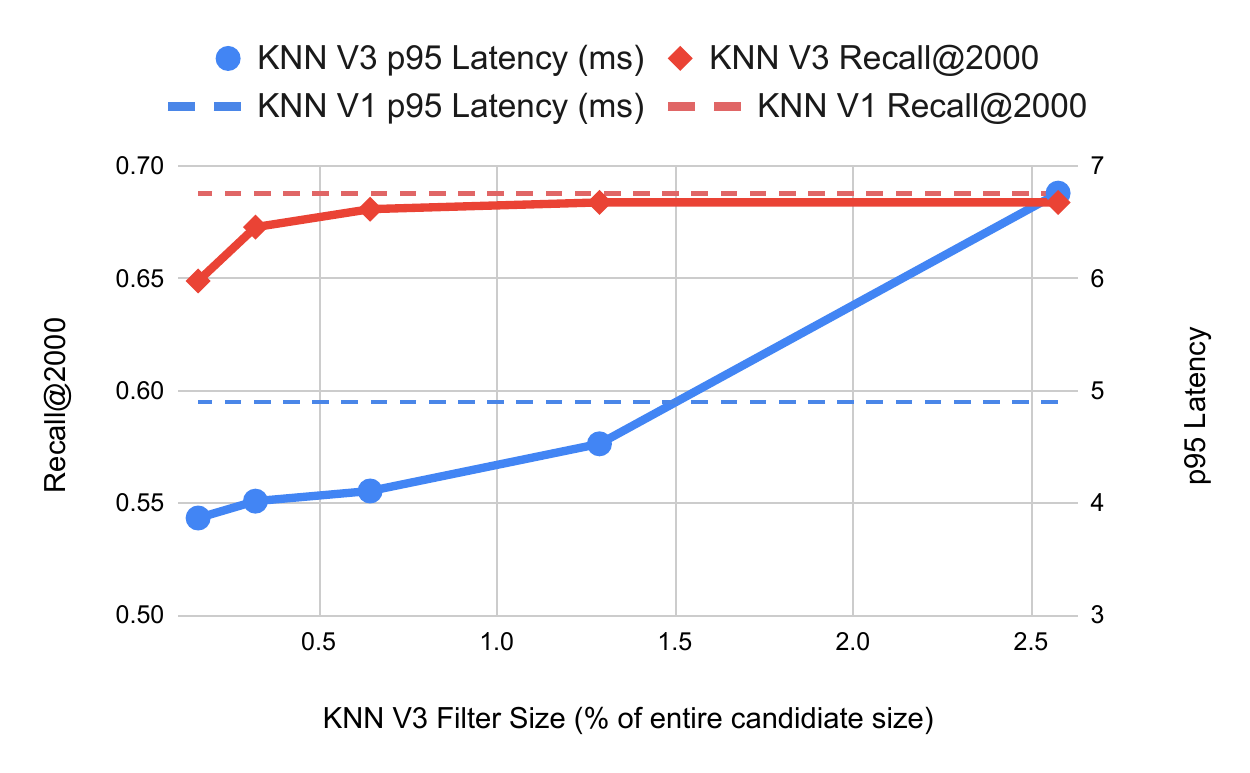}
    \end{adjustbox}
    \caption{\small Trade-off of latency and recall correlated with the filter size of V3 quantized KNN with ABM on high-pass-rate dataset.}
    \label{fig:high-pass-v3-tradeoff}
    \vspace{-12pt}
\end{figure}

From Table~\ref{table:high_pass_knn}, we see that on the high-pass-rate dataset, both TF and PyTorch implementations of V1 (exhaustive search) are faster than V2 (explicit pre-filtering). This is likely due to the native slicing and copying operations in TF and PyTorch, which are especially slow for large matrices, as in V2 with high-pass-rate filters. Benchmarking individual operations revealed that the top-K selection in the latest TF version is slower than in PyTorch, while large-matrix slicing is slower in PyTorch than in TF, leading to performance differences between frameworks. V1's implementation in the high-pass-rate dataset benefits more from the PyTorch implementation with increased batch sizes. Testing the V3 quantized KNN version showed further latency improvements with a trade-off in recall. In this experiment, we used 512-bit quantized embeddings and explored filtering different percentages of items based on quantized embedding similarity before full-precision similarity calculation. The trade-off between latency and recall, correlated with the filter size hyperparameter, is shown in Figure~\ref{fig:high-pass-v3-tradeoff}. By retaining 1\% of items with an additional approximate ranking stage, we achieved around 10\% further latency improvement with nearly parity performance.

\begin{table}[ht]
\centering
\small
\vspace{-4pt}
%\scriptsize % This command makes the text of the table smaller
\begin{tabular}{lcccc}
\hline
\textbf{Method} & \textbf{Batch} & 
\textbf{\begin{tabular}[c]{@{}c@{}}Avg. Latency\\ (ms/batch)\end{tabular}} &

\textbf{\begin{tabular}[c]{@{}c@{}}P95 Latency\\ (ms/batch)\end{tabular}}
%\textbf{Avg. Latency (ms/batch)}  & \textbf{P95 Latency (ms/batch)} 

\\\hline %& \textbf{Recall label@2k} \\ \hline
TF-V2              & 1 & 3.4   & 4.5  \\ % & 100\% \\
PyTorch-V2         & 1 & 1.9  & 2.1  \\\hline % & 100\% \\\hline
% CUDA-V2            & 1 &   &   \\\hline % & 100\% \\\hline
TF-V2              & 16 & 14.2  & 14.8 \\ % 100\% \\
PyTorch-V2         & 16 & 21.4  & 21.9 \\\hline % 100\% \\\hline
% CUDA-V2            & 16 &    &  & 100\% \\\hline
\end{tabular}
\caption{\small Comparison of implementations on low-pass-rate dataset.}
\label{table:low_pass_knn}
\vspace{-16pt}
\end{table}

%From Table~\ref{table:high_pass_knn}, we can see that, on high-pass-rate dataset, the Both TF and PyTorch implementation of V1 exhaustive search is faster than the V2. We attribute this to the fact that the native slicing and copying operations in TF and PyTorch is especially for large matrices, which is the case in the V2 for high-pass-rate filters. Also, via benchmarking different operations individually, we found that the top-K selection operation in the latest TF version is also slower than the PyTorch implementation while the large-matrix slicing operation is slower in PyTorch than TF, leading to the difference in same KNN variant implemented via the two frameworks. Given the benefit of V1 implementation in high-pass-rate dataset, we can gain larger benefits from the PyTorch implementation by increasing the batch size. 

%We also tested the V3 quantized KNN version and saw further latency improvement with a trade-off on the recall. We generate 512 bits quantized embeddings in this experiment and explore different percentage of filtered items based based on the quantize embedding similarity before full-precision similarity calculation. The trade-off between the latency and the recall correlated with the filter size hyperparameter selection is shown in Figure~\ref{fig:high-pass-v3-tradeoff}. By keeping 1\% of items with an extra approximated ranking stage, we could got around 10\% further latency improvement with roughly parity performance.

\subsubsection{Low-Pass-Rate ABM Dataset Benchmarking}\label{sec:low-pass}
%Since V2 version excels in low-pass-rate datasets compared to the other versions (V3 may introduce redundant quantized matrix computation and filtering in this scenario), we compare its performance implemented with TF and PyTorch in Table~\ref{table:low_pass_knn}. For a single query, PyTorch shows an advantage, but TF performs better with larger batch sizes. This is likely due to TF's superior parallel schema for conducting retrievals in parallel. Although queries are fetched in batches, each query has different filters, leading to varied sets and numbers of retrieved items. Therefore, we split the query batch and perform retrieval independently in parallel. Given that V2 is an exhaustive KNN search without liquidity issues, there is no recall drop, and the results are reported here.
As V2 version has the superiority on the low-pass-rate dataset compared to the other two versions (V3 may introduce redundant quantize matrix computation and filtering in the low-pass-rate case), we compare the its performance implemented with TF and PyTorch in Table~\ref{table:low_pass_knn}. One single query, PyTorch still shows its advantage, but TF performs better on larger batch size. We attribute this to the fact the TF has better parallel schema for our case to conduct the retrieval in parallel. Though the queries are fetched in batch, since each query has different filters leading to different sets and number of retrieved items, we split the query batch and conduct the retrieval in parallel for each query independently. Considering that V2 is an exhaustive KNN search without liquidity issue, no recall drop and results are reported here.

\begin{table}[ht]
\centering
\small
\vspace{-4pt}
%\scriptsize % This command makes the text of the table smaller
\begin{tabular}{@{}lccccc@{}}
    \hline
\textbf{Update Per Sec} & \textbf{Batch} & \textbf{QPS} & 
\textbf{\begin{tabular}[c]{@{}c@{}}Avg. Latency\\ (ms/batch)\end{tabular}} &

\textbf{\begin{tabular}[c]{@{}c@{}}P95 Latency\\ (ms/batch)\end{tabular}}
%\textbf{Avg. Latency (ms/batch)}  
%& \textbf{P95 Latency (ms/batch)
\\\hline %& \textbf{Recall label@2k} \\ \hline

0              & 1 & 218 & 4.57  &  4.79 \\ % & 100\% \\
300         & 1 & 215 & 4.64  &  4.93 \\ % & 100\% \\
600         & 1 & 217 & 4.58  &  4.80 \\\hline % & 100\% \\\hline
0              & 5 & 93 & 10.66  &  11.10 \\ % & 100\% \\
300         & 5 & 93 & 10.70  &  11.15 \\ % & 100\% \\\hline
600         & 5 & 93 & 10.70  &  11.16 \\\hline % & 100\% \\\hline

\end{tabular}
\caption{\small Inference latency with concurrent model update.}
\label{table:low_pass_knn}
\vspace{-12pt}
\end{table}

% \begin{table}[ht]
% \centering
% \small % This command makes the text of the table smaller
% \begin{tabular}{lcccc}
% \hline
% \textbf{Method} & \textbf{Batch} & \textbf{Avg. Latency (ms)}  & \textbf{P95 Latency (ms)} \\\hline %& \textbf{Recall label@2k} \\ \hline
% TF-V2              & 1 & 4.0   &  4.7 \\ % & 100\% \\
% PyTorch-V2         & 1 & 2.8   &  3.1 \\\hline % & 100\% \\\hline
% % CUDA-V2            & 1 & 2.5   &  2.6 \\\hline % & 100\% \\\hline
% TF-V2              & 16 &  23.1  & 24.7 & \\ % 100\% \\
% PyTorch-V2         & 16 &  39.2  & 37.2 & \\\hline % 100\% \\\hline
% % CUDA-V2            & 16 &  23.5  & 24.0 & 100\% \\\hline
% \end{tabular}
% \caption{Comparison of different implementations on low-pass-rate dataset on single V100 GPU.}
% \label{table:low_pass_knn}
% \end{table}

% \subsubsection{Stretch Test on Single A100 GPU}
We also conduct a stretch testing on single A100 GPU to measure the capacity of the exhaustive search method on handling large amount of items. For plain KNN with ABM (V1 \& V2) on the high-pass-rate dataset, we are able to handle upto 240 million embeddings with 128 dim and fp16 precision for top-2k selection with single query. For the quantized KNN on an internal notification use case, which we select top-50million members from 1 billion members (64 dimensional embedding saved as fp16) to send relevant notifications, the 1-bit quantized KNN method with 64 bits quantized embedding size can reduce the original 120GB embedding memory to 7.5GB. When processing single query on an A100 GPU, it achieves maximum 21GB high-bandwidth memory with 97.6ms p95 latency.

\subsubsection{Impact of Live Model Update on Inference}\label{sec:low-pass}
We run a benchmark to measure the impact of live model update on the inference latency on a single A100 GPU using our native serving system and bench marking tool. The bench marking tool uses a client for the native serving service and issues requests serially. We use plain KNN model with ABM (V1) and repeat the runs with various concurrent update rates and request batch sizes. We observe no measurable impact on the latency with increased update rate.

\section{Deployment lessons}\label{sec:deployment_lessons}
%We started deployment of {\systemname} with offline inference and observed that we are loosing on some of the fresh candidates, in particular we in A/B test we noticed that live updates plays a key role to serve freshly created posts at LinkedIn. We observed that live-updates can bring relative gains of +6\% in our production systems when live-update is enabled. That is why live updates are so important for improved system performance.

\textit{Freshness:} We initially deployed {\systemname} with offline inference and found it missed some fresh candidates. A/B tests revealed that live updates are crucial for serving newly created LinkedIn posts. Enabling live updates resulted in a +6\% gain in our production systems, highlighting their importance for improved performance.

\textit{Pre-filtering}: Our system employs EBR with pre-filtering, significantly enhancing retrieval quality. Many existing EBR infrastructures use KNN search with post-filtering, where results are first retrieved by KNN distance and then filtered. This post-filtering approach reduces system recall and quality by wasting candidate slots on items that don't meet attribute constraints. By enabling pre-filtering on GPU retrieval, we greatly improved the quality of results compared to our production FAISS and lucene-based systems.

\textit{Custom filtering kernel}: One lesson we learned early is that native TF or PyTorch do not effectively support filtering operations because deep learning frameworks weren't initially designed for model-based retrieval indexes. Native boolean masking and indexing cause a 100X latency increase, making them impractical for production. Therefore, we implemented a custom CUDA solution for pre-filtering on the GPU, which scans items in memory to find those that meet constraints. One approach is to create a CUDA filtering kernel and fuse it with the matrix multiplication kernel to perform masked-matrix multiplication for KNN with pre-filtering. However, this solution is hard to generalize to other similarity measures or operations, as each new architecture would require re-implementation and fine-tuning, slowing down model development and deployment. 
In practice, we could make a trade-off of fully fused kernels and separately implemented kernels. For products needing regular KNN support, fusing the entire kernel with top-k selection and quantization improves serving speed. For general use cases, we create individual custom operations, like pre-filtering and quantization, to allow flexible development and deployment of advanced selection strategy with native neural network operations supported by TF and PyTorch. It is noteworthy that all the results reported in the paper are based on the second solution (even for the three plain KNN version) without extra custom kernel fusion for the purpose of self-consistency and generalizability.

\section{Conclusion}\label{sec:conclusion}

In this paper, we introduced {\systemname}, a state-of-the-art model-based embedding retrieval solution for LinkedIn's production system. Deploying {\systemname} to our online systems resulted in significant improvements in Out-Of-Network post recommendations on the LinkedIn Feed. We believe we are among the first in the industry to support live-updated, differentiable model-based indexing for recommendation and search applications. Looking forward, {\systemname} paves the way for unifying retrieval and ranking into a single GPU model, simplifying complex infrastructure and allowing end-to-end optimization of the entire differentiable system with gradient descent.

\section{Acknowledgements}\label{sec:acknowledgements}
The authors would like to thank Jerry Shen, Yuchin Juan, Xiaobing Xue, Souvik Ghosh, Amol Ghoting, Vivek Hariharan, Ping Li, Luke Simon and others who collaborated with us.

\balance

\bibliographystyle{ACM-Reference-Format}
\bibliography{bibliography}


\begin{thebibliography}{25}

%%% the Web address.

\ifx \showCODEN    \undefined \def \showCODEN     #1{\unskip}     \fi
\ifx \showDOI      \undefined \def \showDOI       #1{#1}\fi
\ifx \showISBNx    \undefined \def \showISBNx     #1{\unskip}     \fi
\ifx \showISBNxiii \undefined \def \showISBNxiii  #1{\unskip}     \fi
\ifx \showISSN     \undefined \def \showISSN      #1{\unskip}     \fi
\ifx \showLCCN     \undefined \def \showLCCN      #1{\unskip}     \fi
\ifx \shownote     \undefined \def \shownote      #1{#1}          \fi
\ifx \showarticletitle \undefined \def \showarticletitle #1{#1}   \fi
\ifx \showURL      \undefined \def \showURL       {\relax}        \fi
% invisible to TeX
\providecommand\bibfield[2]{#2}
\providecommand\bibinfo[2]{#2}
\providecommand\natexlab[1]{#1}
\providecommand\showeprint[2][]{arXiv:#2}

\bibitem[Borisyuk et~al\mbox{.}(2024)]%
        {borisyuk2024lignn}
\bibfield{author}{\bibinfo{person}{Fedor Borisyuk}, \bibinfo{person}{Shihai He}, \bibinfo{person}{Yunbo Ouyang}, \bibinfo{person}{Morteza Ramezani}, \bibinfo{person}{Peng Du}, \bibinfo{person}{Xiaochen Hou}, \bibinfo{person}{Chengming Jiang}, \bibinfo{person}{Nitin Pasumarthy}, \bibinfo{person}{Priya Bannur}, \bibinfo{person}{Birjodh Tiwana}, \bibinfo{person}{Ping Liu}, \bibinfo{person}{Siddharth Dangi}, \bibinfo{person}{Daqi Sun}, \bibinfo{person}{Zhoutao Pei}, \bibinfo{person}{Xiao Shi}, \bibinfo{person}{Sirou Zhu}, \bibinfo{person}{Qianqi Shen}, \bibinfo{person}{Kuang-Hsuan Lee}, \bibinfo{person}{David Stein}, \bibinfo{person}{Baolei Li}, \bibinfo{person}{Haichao Wei}, \bibinfo{person}{Amol Ghoting}, {and} \bibinfo{person}{Souvik Ghosh}.} \bibinfo{year}{2024}\natexlab{}.
\newblock \showarticletitle{LiGNN: Graph Neural Networks at LinkedIn}. In \bibinfo{booktitle}{\emph{KDD}}.
\newblock


\bibitem[Chen et~al\mbox{.}(2023)]%
        {Lucene_chen2023endtoend}
\bibfield{author}{\bibinfo{person}{Haonan Chen}, \bibinfo{person}{Carlos Lassance}, {and} \bibinfo{person}{Jimmy Lin}.} \bibinfo{year}{2023}\natexlab{}.
\newblock \bibinfo{title}{End-to-End Retrieval with Learned Dense and Sparse Representations Using Lucene}.
\newblock
\newblock
\showeprint[arxiv]{2311.18503}~[cs.IR]


\bibitem[Curtiss et~al\mbox{.}(2013)]%
        {unicorn_paper}
\bibfield{author}{\bibinfo{person}{Michael Curtiss}, \bibinfo{person}{Iain Becker}, \bibinfo{person}{Tudor Bosman}, \bibinfo{person}{Sergey Doroshenko}, \bibinfo{person}{Lucian Grijincu}, \bibinfo{person}{Tom Jackson}, \bibinfo{person}{Sandhya Kunnatur}, \bibinfo{person}{Soren Lassen}, \bibinfo{person}{Philip Pronin}, \bibinfo{person}{Sriram Sankar}, \bibinfo{person}{Guanghao Shen}, \bibinfo{person}{Gintaras Woss}, \bibinfo{person}{Chao Yang}, {and} \bibinfo{person}{Ning Zhang}.} \bibinfo{year}{2013}\natexlab{}.
\newblock \showarticletitle{Unicorn: A System for Searching the Social Graph}. In \bibinfo{booktitle}{\emph{VLDB}}.
\newblock


\bibitem[Guo et~al\mbox{.}(2020)]%
        {scann_google_paper}
\bibfield{author}{\bibinfo{person}{Ruiqi Guo}, \bibinfo{person}{Philip Sun}, \bibinfo{person}{Erik Lindgren}, \bibinfo{person}{Quan Geng}, \bibinfo{person}{David Simcha}, \bibinfo{person}{Felix Chern}, {and} \bibinfo{person}{Sanjiv Kumar}.} \bibinfo{year}{2020}\natexlab{}.
\newblock \showarticletitle{Accelerating Large-Scale Inference with Anisotropic Vector Quantization}. \bibinfo{publisher}{ICML}.
\newblock


\bibitem[GV(2008)]%
        {veniceDB}
\bibfield{author}{\bibinfo{person}{Félix GV}.} \bibinfo{year}{2008}\natexlab{}.
\newblock \bibinfo{title}{Open Sourcing Venice – LinkedIn’s Derived Data Platform}.
\newblock
\newblock
\newblock
\shownote{https://www.linkedin.com/blog/engineering/open-source/open-sourcing-venice-linkedin-s-derived-data-platform}.


\bibitem[Huang et~al\mbox{.}(2020)]%
        {fb_search_embeddings_paper}
\bibfield{author}{\bibinfo{person}{Jui-Ting Huang}, \bibinfo{person}{Ashish Sharma}, \bibinfo{person}{Shuying Sun}, \bibinfo{person}{Li Xia}, \bibinfo{person}{David Zhang}, \bibinfo{person}{Philip Pronin}, \bibinfo{person}{Janani Padmanabhan}, \bibinfo{person}{Giuseppe Ottaviano}, {and} \bibinfo{person}{Linjun Yang}.} \bibinfo{year}{2020}\natexlab{}.
\newblock \showarticletitle{Embedding-Based Retrieval in Facebook Search}. In \bibinfo{booktitle}{\emph{KDD}}.
\newblock


\bibitem[Jiang et~al\mbox{.}(2019)]%
        {XDL_paper}
\bibfield{author}{\bibinfo{person}{Biye Jiang}, \bibinfo{person}{Chao Deng}, \bibinfo{person}{Huimin Yi}, \bibinfo{person}{Zelin Hu}, \bibinfo{person}{Guorui Zhou}, \bibinfo{person}{Yang Zheng}, \bibinfo{person}{Sui Huang}, \bibinfo{person}{Xinyang Guo}, \bibinfo{person}{Dongyue Wang}, \bibinfo{person}{Yue Song}, \bibinfo{person}{Liqin Zhao}, \bibinfo{person}{Zhi Wang}, \bibinfo{person}{Peng Sun}, \bibinfo{person}{Yu Zhang}, \bibinfo{person}{Di Zhang}, \bibinfo{person}{Jinhui Li}, \bibinfo{person}{Jian Xu}, \bibinfo{person}{Xiaoqiang Zhu}, {and} \bibinfo{person}{Kun Gai}.} \bibinfo{year}{2019}\natexlab{}.
\newblock \showarticletitle{XDL: An Industrial Deep Learning Framework for High-Dimensional Sparse Data}. \bibinfo{publisher}{KDD}.
\newblock


\bibitem[Johnson et~al\mbox{.}(2021)]%
        {FAISS_paper}
\bibfield{author}{\bibinfo{person}{Jeff Johnson}, \bibinfo{person}{Matthijs Douze}, {and} \bibinfo{person}{Hervé Jégou}.} \bibinfo{year}{2021}\natexlab{}.
\newblock \showarticletitle{Billion-Scale Similarity Search with GPUs}.
\newblock \bibinfo{journal}{\emph{IEEE Transactions on Big Data}} (\bibinfo{year}{2021}).
\newblock


\bibitem[Jégou et~al\mbox{.}(2011)]%
        {pq_paper}
\bibfield{author}{\bibinfo{person}{Herve Jégou}, \bibinfo{person}{Matthijs Douze}, {and} \bibinfo{person}{Cordelia Schmid}.} \bibinfo{year}{2011}\natexlab{}.
\newblock \showarticletitle{Product Quantization for Nearest Neighbor Search}.
\newblock \bibinfo{journal}{\emph{IEEE Transactions on Pattern Analysis and Machine Intelligence}} (\bibinfo{year}{2011}).
\newblock


\bibitem[Li and Li(2023)]%
        {li2023oporp}
\bibfield{author}{\bibinfo{person}{Ping Li} {and} \bibinfo{person}{Xiaoyun Li}.} \bibinfo{year}{2023}\natexlab{}.
\newblock \showarticletitle{OPORP: One permutation+ one random projection}.
\newblock \bibinfo{journal}{\emph{arXiv preprint arXiv:2302.03505}} (\bibinfo{year}{2023}).
\newblock


\bibitem[Lian et~al\mbox{.}(2022)]%
        {persia_paper}
\bibfield{author}{\bibinfo{person}{Xiangru Lian}, \bibinfo{person}{Binhang Yuan}, \bibinfo{person}{Xuefeng Zhu}, \bibinfo{person}{Yulong Wang}, \bibinfo{person}{Yongjun He}, \bibinfo{person}{Honghuan Wu}, \bibinfo{person}{Lei Sun}, \bibinfo{person}{Haodong Lyu}, \bibinfo{person}{Chengjun Liu}, \bibinfo{person}{Xing Dong}, \bibinfo{person}{Yiqiao Liao}, \bibinfo{person}{Mingnan Luo}, \bibinfo{person}{Congfei Zhang}, \bibinfo{person}{Jingru Xie}, \bibinfo{person}{Haonan Li}, \bibinfo{person}{Lei Chen}, \bibinfo{person}{Renjie Huang}, \bibinfo{person}{Jianying Lin}, \bibinfo{person}{Chengchun Shu}, \bibinfo{person}{Xuezhong Qiu}, \bibinfo{person}{Zhishan Liu}, \bibinfo{person}{Dongying Kong}, \bibinfo{person}{Lei Yuan}, \bibinfo{person}{Hai Yu}, \bibinfo{person}{Sen Yang}, \bibinfo{person}{Ce Zhang}, {and} \bibinfo{person}{Ji Liu}.} \bibinfo{year}{2022}\natexlab{}.
\newblock \showarticletitle{Persia: An Open, Hybrid System Scaling Deep Learning-Based Recommenders up to 100 Trillion Parameters}. In \bibinfo{booktitle}{\emph{KDD}}.
\newblock


\bibitem[Liu et~al\mbox{.}(2021)]%
        {Que2Search_paper}
\bibfield{author}{\bibinfo{person}{Yiqun Liu}, \bibinfo{person}{Kaushik Rangadurai}, \bibinfo{person}{Yunzhong He}, \bibinfo{person}{Siddarth Malreddy}, \bibinfo{person}{Xunlong Gui}, \bibinfo{person}{Xiaoyi Liu}, {and} \bibinfo{person}{Fedor Borisyuk}.} \bibinfo{year}{2021}\natexlab{}.
\newblock \showarticletitle{Que2Search: Fast and Accurate Query and Document Understanding for Search at Facebook}. \bibinfo{publisher}{KDD}.
\newblock


\bibitem[Liu et~al\mbox{.}(2022)]%
        {liu2022monolith}
\bibfield{author}{\bibinfo{person}{Zhuoran Liu}, \bibinfo{person}{Leqi Zou}, \bibinfo{person}{Xuan Zou}, \bibinfo{person}{Caihua Wang}, \bibinfo{person}{Biao Zhang}, \bibinfo{person}{Da Tang}, \bibinfo{person}{Bolin Zhu}, \bibinfo{person}{Yijie Zhu}, \bibinfo{person}{Peng Wu}, \bibinfo{person}{Ke Wang}, {and} \bibinfo{person}{Youlong Cheng}.} \bibinfo{year}{2022}\natexlab{}.
\newblock \bibinfo{title}{Monolith: Real Time Recommendation System With Collisionless Embedding Table}.
\newblock
\newblock
\showeprint[arxiv]{2209.07663}~[cs.IR]


\bibitem[Malkov and Yashunin(2020)]%
        {hnsw_paper}
\bibfield{author}{\bibinfo{person}{Yu~A. Malkov} {and} \bibinfo{person}{D.~A. Yashunin}.} \bibinfo{year}{2020}\natexlab{}.
\newblock \showarticletitle{Efficient and Robust Approximate Nearest Neighbor Search Using Hierarchical Navigable Small World Graphs}.
\newblock \bibinfo{journal}{\emph{IEEE Trans. Pattern Anal. Mach. Intell.}} (\bibinfo{year}{2020}).
\newblock


\bibitem[Nolet(2023)]%
        {RAFT_paper}
\bibfield{author}{\bibinfo{person}{Corey Nolet}.} \bibinfo{year}{2023}\natexlab{}.
\newblock \bibinfo{booktitle}{\emph{Reusable Computational Patterns for Machine Learning and Information Retrieval with RAPIDS RAFT}}.
\newblock
\urldef\tempurl%
\url{https://developer.nvidia.com/blog/reusable-computational-patterns-for-machine-learning-and-data-analytics-with-rapids-raft/}
\showURL{%
\tempurl}


\bibitem[Ootomo et~al\mbox{.}(2023)]%
        {ootomo2023cagra}
\bibfield{author}{\bibinfo{person}{Hiroyuki Ootomo}, \bibinfo{person}{Akira Naruse}, \bibinfo{person}{Corey Nolet}, \bibinfo{person}{Ray Wang}, \bibinfo{person}{Tamas Feher}, {and} \bibinfo{person}{Yong Wang}.} \bibinfo{year}{2023}\natexlab{}.
\newblock \bibinfo{title}{CAGRA: Highly Parallel Graph Construction and Approximate Nearest Neighbor Search for GPUs}.
\newblock
\newblock
\showeprint[arxiv]{2308.15136}~[cs.DS]


\bibitem[Pal et~al\mbox{.}(2020)]%
        {PinnerSage_paper}
\bibfield{author}{\bibinfo{person}{Aditya Pal}, \bibinfo{person}{Chantat Eksombatchai}, \bibinfo{person}{Yitong Zhou}, \bibinfo{person}{Bo Zhao}, \bibinfo{person}{Charles Rosenberg}, {and} \bibinfo{person}{Jure Leskovec}.} \bibinfo{year}{2020}\natexlab{}.
\newblock \showarticletitle{PinnerSage: Multi-Modal User Embedding Framework for Recommendations at Pinterest}. \bibinfo{publisher}{KDD}.
\newblock


\bibitem[Rajput et~al\mbox{.}(2023)]%
        {rajput2023recommender}
\bibfield{author}{\bibinfo{person}{Shashank Rajput}, \bibinfo{person}{Nikhil Mehta}, \bibinfo{person}{Anima Singh}, \bibinfo{person}{Raghunandan~H. Keshavan}, \bibinfo{person}{Trung Vu}, \bibinfo{person}{Lukasz Heldt}, \bibinfo{person}{Lichan Hong}, \bibinfo{person}{Yi Tay}, \bibinfo{person}{Vinh~Q. Tran}, \bibinfo{person}{Jonah Samost}, \bibinfo{person}{Maciej Kula}, \bibinfo{person}{Ed~H. Chi}, {and} \bibinfo{person}{Maheswaran Sathiamoorthy}.} \bibinfo{year}{2023}\natexlab{}.
\newblock \bibinfo{title}{Recommender Systems with Generative Retrieval}.
\newblock
\newblock
\showeprint[arxiv]{2305.05065}~[cs.IR]


\bibitem[Rangadurai et~al\mbox{.}(2022)]%
        {nxtpost_paper}
\bibfield{author}{\bibinfo{person}{Kaushik Rangadurai}, \bibinfo{person}{Yiqun Liu}, \bibinfo{person}{Siddarth Malreddy}, \bibinfo{person}{Xiaoyi Liu}, \bibinfo{person}{Piyush Maheshwari}, \bibinfo{person}{Vishwanath Sangale}, {and} \bibinfo{person}{Fedor Borisyuk}.} \bibinfo{year}{2022}\natexlab{}.
\newblock \showarticletitle{NxtPost: User To Post Recommendations In Facebook Groups}. In \bibinfo{booktitle}{\emph{KDD}}.
\newblock


\bibitem[Shen et~al\mbox{.}(2024)]%
        {shen2024learningretrievejobmatching}
\bibfield{author}{\bibinfo{person}{Jianqiang Shen}, \bibinfo{person}{Yuchin Juan}, \bibinfo{person}{Shaobo Zhang}, \bibinfo{person}{Ping Liu}, \bibinfo{person}{Wen Pu}, \bibinfo{person}{Sriram Vasudevan}, \bibinfo{person}{Qingquan Song}, \bibinfo{person}{Fedor Borisyuk}, \bibinfo{person}{Kay~Qianqi Shen}, \bibinfo{person}{Haichao Wei}, \bibinfo{person}{Yunxiang Ren}, \bibinfo{person}{Yeou~S. Chiou}, \bibinfo{person}{Sicong Kuang}, \bibinfo{person}{Yuan Yin}, \bibinfo{person}{Ben Zheng}, \bibinfo{person}{Muchen Wu}, \bibinfo{person}{Shaghayegh Gharghabi}, \bibinfo{person}{Xiaoqing Wang}, \bibinfo{person}{Huichao Xue}, \bibinfo{person}{Qi Guo}, \bibinfo{person}{Daniel Hewlett}, \bibinfo{person}{Luke Simon}, \bibinfo{person}{Liangjie Hong}, {and} \bibinfo{person}{Wenjing Zhang}.} \bibinfo{year}{2024}\natexlab{}.
\newblock \bibinfo{title}{Learning to Retrieve for Job Matching}.
\newblock
\newblock
\showeprint[arxiv]{2402.13435}~[cs.IR]
\urldef\tempurl%
\url{https://arxiv.org/abs/2402.13435}
\showURL{%
\tempurl}


\bibitem[Tay et~al\mbox{.}(2022)]%
        {tay2022transformer}
\bibfield{author}{\bibinfo{person}{Yi Tay}, \bibinfo{person}{Vinh~Q. Tran}, \bibinfo{person}{Mostafa Dehghani}, \bibinfo{person}{Jianmo Ni}, \bibinfo{person}{Dara Bahri}, \bibinfo{person}{Harsh Mehta}, \bibinfo{person}{Zhen Qin}, \bibinfo{person}{Kai Hui}, \bibinfo{person}{Zhe Zhao}, \bibinfo{person}{Jai Gupta}, \bibinfo{person}{Tal Schuster}, \bibinfo{person}{William~W. Cohen}, {and} \bibinfo{person}{Donald Metzler}.} \bibinfo{year}{2022}\natexlab{}.
\newblock \showarticletitle{Transformer Memory as a Differentiable Search Index}. \bibinfo{publisher}{NEURIPS}.
\newblock


\bibitem[Zhai et~al\mbox{.}(2023)]%
        {MoL_paper}
\bibfield{author}{\bibinfo{person}{Jiaqi Zhai}, \bibinfo{person}{Zhaojie Gong}, \bibinfo{person}{Yueming Wang}, \bibinfo{person}{Xiao Sun}, \bibinfo{person}{Zheng Yan}, \bibinfo{person}{Fu Li}, {and} \bibinfo{person}{Xing Liu}.} \bibinfo{year}{2023}\natexlab{}.
\newblock \showarticletitle{Revisiting Neural Retrieval on Accelerators}. In \bibinfo{booktitle}{\emph{Proceedings of the 29th ACM SIGKDD Conference on Knowledge Discovery and Data Mining}} \emph{(\bibinfo{series}{KDD '23})}. \bibinfo{publisher}{Association for Computing Machinery}, \bibinfo{address}{New York, NY, USA}, \bibinfo{pages}{5520–5531}.
\newblock
\showISBNx{9798400701030}
\urldef\tempurl%
\url{https://doi.org/10.1145/3580305.3599897}
\showDOI{\tempurl}


\bibitem[Zhang et~al\mbox{.}(2023)]%
        {zhang2023modelenhanced}
\bibfield{author}{\bibinfo{person}{Hailin Zhang}, \bibinfo{person}{Yujing Wang}, \bibinfo{person}{Qi Chen}, \bibinfo{person}{Ruiheng Chang}, \bibinfo{person}{Ting Zhang}, \bibinfo{person}{Ziming Miao}, \bibinfo{person}{Yingyan Hou}, \bibinfo{person}{Yang Ding}, \bibinfo{person}{Xupeng Miao}, \bibinfo{person}{Haonan Wang}, \bibinfo{person}{Bochen Pang}, \bibinfo{person}{Yuefeng Zhan}, \bibinfo{person}{Hao Sun}, \bibinfo{person}{Weiwei Deng}, \bibinfo{person}{Qi Zhang}, \bibinfo{person}{Fan Yang}, \bibinfo{person}{Xing Xie}, \bibinfo{person}{Mao Yang}, {and} \bibinfo{person}{Bin Cui}.} \bibinfo{year}{2023}\natexlab{}.
\newblock \bibinfo{title}{Model-enhanced Vector Index}.
\newblock
\newblock


\bibitem[Zhao et~al\mbox{.}(2020)]%
        {song_paper}
\bibfield{author}{\bibinfo{person}{Weijie Zhao}, \bibinfo{person}{Shulong Tan}, {and} \bibinfo{person}{Ping Li}.} \bibinfo{year}{2020}\natexlab{}.
\newblock \showarticletitle{SONG: Approximate Nearest Neighbor Search on GPU}. In \bibinfo{booktitle}{\emph{ICDE}}.
\newblock


\bibitem[Zhao et~al\mbox{.}(2022)]%
        {zhao2022constrained}
\bibfield{author}{\bibinfo{person}{Weijie Zhao}, \bibinfo{person}{Shulong Tan}, {and} \bibinfo{person}{Ping Li}.} \bibinfo{year}{2022}\natexlab{}.
\newblock \showarticletitle{Constrained Approximate Similarity Search on Proximity Graph}.
\newblock \bibinfo{journal}{\emph{arXiv preprint arXiv:2210.14958}} (\bibinfo{year}{2022}).
\newblock


\end{thebibliography}

\end{document}